%% file: main.tex
\documentclass[draft]{agujournal2019}
\usepackage{url} 
\usepackage{lineno}
\usepackage[inline]{trackchanges} 
\usepackage{soul}
\usepackage{amsmath,amsfonts,amsthm}
\graphicspath{{plot/}}
\usepackage{setspace}
\usepackage{subcaption}

\draftfalse

\journalname{Enter journal name here}

\begin{document}

%
%


\title{GeoFUSE: A High-Efficiency Surrogate Model for Seawater Intrusion Prediction and Uncertainty Reduction}

%
%




\authors{Su Jiang\affil{1}, Chuyang Liu\affil{1}, Dipankar Dwivedi\affil{1}}


\affiliation{1}{Lawrence Berkeley National Laboratory}




\correspondingauthor{Su Jiang}{sujiang@lbl.gov}



\begin{keypoints}
\item A new deep-learning framework GeoFUSE is developed for rapid and accurate seawater intrusion simulations and data assimilation 
\item U-Net Fourier Neural Operator (U-FNO) acts as surrogate model to reduce simulation time from hours to seconds, achieve a speedup of 360,000 times
\item An ensemble method, conditioned on observations, is used to reduce uncertainty in geomodels and flow predictions
\end{keypoints}

\section*{Plain Language Summary}
Seawater intrusion is an increasing threat to coastal areas due to climate change and rising sea levels. Predicting the movement of seawater through underground aquifers is essential for managing coastal water resources. However, traditional simulations are often slow, computationally expensive, and susceptible to inaccuracies due to uncertainty in geological systems. We developed a deep-learning-based framework, GeoFUSE, for rapid and accurate subsurface flow simulations that also reduces uncertainty by incorporating measurement data. Using a U-FNO surrogate model, we significantly accelerate the simulation process, while an ensemble method helps further reduce uncertainty in salinity distribution predictions. We demonstrate the effectiveness of the framework using the Beaver Creek tidal stream-floodplain system.

%
%

%
%

\begin{abstract}
Seawater intrusion into coastal aquifers poses a significant threat to groundwater resources, especially with rising sea levels due to climate change. Accurate modeling and uncertainty quantification of this process are crucial but are often hindered by the high computational costs of traditional numerical simulations. In this work, we develop GeoFUSE, a novel deep-learning-based surrogate framework that integrates the U-Net Fourier Neural Operator (U-FNO) with Principal Component Analysis (PCA) and Ensemble Smoother with Multiple Data Assimilation (ESMDA). GeoFUSE enables fast and efficient simulation of seawater intrusion while significantly reducing uncertainty in model predictions. We apply GeoFUSE to a 2D cross-section of the Beaver Creek tidal stream-floodplain system in Washington State. Using 1,500 geological realizations, we train the U-FNO surrogate model to approximate salinity distribution and accumulation. The U-FNO model successfully reduces the computational time from hours (using PFLOTRAN simulations) to seconds, achieving a speedup of approximately 360,000 times while maintaining high accuracy. By integrating measurement data from monitoring wells, the framework significantly reduces geological uncertainty and improves the predictive accuracy of the salinity distribution over a 20-year period. Our results demonstrate that GeoFUSE improves computational efficiency and provides a robust tool for real-time uncertainty quantification and decision making in groundwater management. Future work will extend GeoFUSE to 3D models and incorporate additional factors such as sea-level rise and extreme weather events, making it applicable to a broader range of coastal and subsurface flow systems.
\end{abstract}

\input{introduction}
\input{method}

\input{results}
\input{discussion}
\input{conclusion}
\section*{Data Availability Statement}
The simulation data can be generated by solving the equations using the input parameters presented in the figures and methodology. Data will be made available upon request. The code used in this study will be shared in a GitHub repository by the time of publication.


\acknowledgments
This work was supported by the Automated Scenario Assessment of Groundwater Table \& Salinity Response to Sea-Level Rise Project at Lawrence Berkeley National Laboratory, funded by the U.S. Department of Defense, Strategic Environmental Research and Development Program (SERDP).


\clearpage
\bibliography{swi}

\end{document}

%% file: introduction.tex
\section{Introduction}
Seawater intrusion into freshwater aquifers, driven by climate change and groundwater exploitation, has become an increasing threat to subsurface systems in coastal regions globally \cite{werner2013seawater, ketabchi2016sea}. The effects of seawater intrusion can be detrimental to both the environment and human activities, including contamination of freshwater and drinking water, disruption of ecosystems, impacts on agriculture, and damage to infrastructure. Efficient and accurate modeling of seawater intrusion is crucial for predicting the movement of the saltwater front and ensuring the sustainability of groundwater resources and ecosystems. 

Researchers have applied multiple simulation techniques to address the growing threat of seawater intrusion and predict its impact on coastal aquifers. Tools such as SEAWAT \cite{langevin2006modflow, langevin2008seawat}, SUTRA \cite{provost2019sutra}, and PFLOTRAN \cite{hammond2014evaluating, yabusaki2020floodplain} are commonly used to simulate variable density groundwater flow and solute transport in both conceptual and real-site models. These simulations require a detailed parameterization of geology and aquifer properties, informed by both experiments and numerical modeling. Furthermore, model calibration and data assimilation are essential to improve predictive accuracy and reduce uncertainties. Typically, model calibration is conducted within a Bayesian framework, conditioned on monitoring data, and requires repeated simulations to quantify uncertainty \cite{arora2011inverse, arora2012uncertainty}. However, forward simulations over long periods are computationally expensive, particularly when accounting for pressure changes caused by daily tidal effects. Furthermore, the inherent uncertainty of aquifer parameters in heterogeneous systems introduces significant variability into flow predictions. Despite advancements in modeling, these processes remain computationally intensive, and the requirement for numerous forward runs limits the efficiency of data assimilation. This work aims to reduce computational costs, enabling faster, more accurate predictions and mitigating uncertainty in geomodels and predictions. 

To address computational challenges in subsurface applications, several traditional surrogate modeling techniques have been developed. These include artificial neural networks for multi-objective optimization \cite{bhattacharjya2009ann}, genetic programming and modular neural networks \cite{sreekanth2010multi}, Evolutionary Polynomial Regression \cite{hussain2015surrogate}, and Gaussian process emulators \cite{rajabi2017uncertainty}. These models often focus on specific quantities within simulation-optimization schemes and may not effectively capture the temporal evolution of state variables, such as the spatial distribution of flow and salinity predictions, across entire simulation domains, particularly in geologically complex areas.

In contrast, recent deep learning approaches have shown significant potential in addressing the challenge of capturing the temporal evolution of state variables, particularly for spatio-temporal flow prediction in subsurface systems. For instance, \citeA{mo2019deep} developed deep autoregressive neural networks for predicting contamination transport, and \citeA{zhou2021markov} used convolutional neural networks to identify contaminant sources. Other advancements include the recurrent residual U-net for two-phase flow problems \cite{tang2020deep}, which has been further adapted for multi-fidelity systems \cite{jiang2023use} and various geological scenarios \cite{han2024surrogate}. The Fourier Neural Operator (FNO) \cite{li2020fourier}, designed to capture fluid dynamics in infinite-dimensional Fourier spaces, and the U-FNO, which combines FNO with a U-net architecture for improved accuracy and efficiency in geological carbon storage \cite{wen2022u}, are notable innovations. U-FNO has been successfully applied to groundwater contamination problems, predicting flow responses under new parameter inputs and boundary conditions \cite{meray2024physics}. While U-FNO, with its strong ability to capture the temporal evolution of state variables, is particularly well-suited for seawater intrusion problems, it stands out for its efficiency in modeling complex interactions between saltwater and freshwater flows. This makes it a highly effective approach for generating accurate predictions under varying conditions, though other methods may also offer viable solutions depending on specific problem requirements and constraints.

To address the uncertainty inherent in heterogeneous subsurface geological systems, inverse modeling and data assimilation, widely applied in seawater intrusion studies, are essential for calibrating geological models based on monitoring data. For instance, \citeA{yoon2017maximizing} used pressure data with the Ensemble Kalman Filter (EnKF) to calibrate heterogeneous models for a saline aquifer, while \citeA{dodangeh2022joint} employed concentration measurements for joint inversion of contaminant sources and hydraulic conductivity in coastal aquifers. Similarly, \citeA{goebel2017resistivity} utilized Electrical Resistivity Tomography for inversion in Monterey's coastal region, and \citeA{cao2024deep} proposed a deep-learning-based data assimilation method to infer non-Gaussian heterogeneous coastal aquifers. Recently, the Ensemble Smoother with Multiple Data Assimilation (ESMDA) method, developed by~\citeA{emerick2013ensemble}, is a variant version of EnKF, has been widely applied to various subsurface flow problems, including contaminant transport \cite{zhou2022deep} and geological carbon storage \cite{jiang2024history}. The ESMDA method has demonstrated its ability to assimilate noisy and sparse measurements and can be extended to calibrate geological models for coastal systems. However, the computational costs of these inversion processes remain high, highlighting the need for integrating surrogate models to improve efficiency.

This work focuses on developing an efficient surrogate model to enable rapid seawater intrusion prediction and integrating it into a data assimilation framework for calibrating heterogeneous geological models. The ultimate goal is to reduce prediction uncertainty while improving computational efficiency. The framework is applied to a 2D cross-section of the Beaver Creek tidal stream-floodplain system \cite{yabusaki2020floodplain}, which serves as a representative case to assess the effects of tidal dynamics and the potential impacts of sea level rise on coastal systems.

The framework combines a surrogate model trained on PFLOTRAN-simulated data with a data assimilation process using Principal Component Analysis (PCA) and the ESMDA. This approach enables parameter reduction, model calibration, and uncertainty reduction in salinity predictions. By incorporating this method, we aim to significantly reduce computational costs while maintaining predictive accuracy, ultimately advancing subsurface modeling for more effective management of seawater intrusion in coastal environments.

%% file: method.tex
\section{Methodology}\label{sec:method}
In this work, we apply U-Fourier neural operator (FNO) for rapid spatio-temporal flow predictions, while ESMDA is integrated conditioned on observational data to calibrate geological models, thereby reducing computational costs in seawater intrusion modeling and improving predictive accuracy for heterogeneous geological models. This combined framework, referred to as GeoFUSE (explained further in Section~\ref{sec:GeoFUSE}), is applied to a 2D cross-section of the Beaver Creek tidal stream-floodplain system \cite{yabusaki2020floodplain}. The U-FNO surrogate is trained on the simulation results of 2D multi-Gaussian permeability realizations. 

The Beaver Creek model was developed by \citeA{yabusaki2020floodplain}, who investigated the effects of tidal restoration on floodplain inundation and salinization by modeling subsurface flow and transport. Their study focused on the dynamics of a first-order tidal stream system, aiming to clarify the interactions between freshwater and seawater, as well as the processes controlling water levels and salinity in the floodplain. The Beaver Creek site is ideal for this study, as it improves our understanding of how tidal influences reshape hydrological and ecological patterns in restored environments. The complexities of tidal effects, salinization, and uncertain geological models make the problem computationally intensive. In the following sections, we describe the Beaver Creek model, surrogate model development, the GeoFUSE data assimilation framework, and performance evaluation in detail.

\subsection{Beaver Creek Model and Governing Equations Using PFLOTRAN}
The Beaver Creek model from \citeA{yabusaki2020floodplain} was developed using PFLOTRAN \cite{hammond2014evaluating}, a computational tool designed to simulate subsurface flow and reactive transport processes in variably saturated porous media. Given these capabilities, PFLOTRAN is particularly useful for studying seawater intrusion as it effectively captures hydrological and biogeochemical dynamics in complex environments such as aquifers and floodplains. By employing advanced numerical techniques to solve flow and transport equations, PFLOTRAN enables researchers to simulate water movement, salinity intrusion, and contaminant transport scenarios.

\citeA{yabusaki2020floodplain} implemented a single-phase solute transport model in variably saturated porous media to simulate the Beaver Creek system. The 2D flow and salinity transport simulations were conducted with the PFLOTRAN simulator \cite{hammond2014evaluating}. Darcy's and Richard's equations were applied to describe 2D flow and transport process, making them well-suited for capturing the dynamics of groundwater flow and salinity intrusion. The governing mass conservation equation is written as
\begin{linenomath*}
\begin{equation}\label{eq:gov_eq}
\frac{\partial}{\partial t}(\phi S \eta) + \nabla \cdot (\eta \mathbf{q}) = Q_w, 
\end{equation}
\end{linenomath*}
where $\phi$ is the porosity [-], $S$ is saturation [-], $\eta$ is the water molar density [kmol/m$^3$], $\mathbf{q}$ represents the water flux [m/s], and $Q_w$ represents the source/sink term [kmol/m$^3$/s]. The Darcy flux $\mathbf{q}$ is defined by
\begin{linenomath*}
\begin{equation}\label{eq:darcy_flow}
\mathbf{q} = - \frac{\mathbf{k}k_r(S)}{\mu}\nabla(P-\rho g z), 
\end{equation}
\end{linenomath*}
where $\mathbf{k}$ denotes the intrinsic permeability tensor [m$^2$], $k_r(S)$ denotes the relative permeability [-] (the relationship between $k_r$ and $S$ is derived on laboratory experiments), $\mu$ is the viscosity [Pa$\cdot$s], $P$ is the pressure [Pa], $\rho$ is the water density [kg/m$^3$], $g$ is the gravitational field [m/s$^2$], and $z$ is the vertical position [m]. The van Genuchten function is used for the water retention and relative permeability curve, with specific parameters such as a residual saturation of 0.15, alpha of $2 \times 10^{-4}$ Pa, and m of 0.2908.

For the seawater intrusion model, the reactive transport equation for salt mass considering advection and diffusion is written as
\begin{linenomath*}
\begin{equation}\label{eq:advec_diff}
\frac{\partial \phi c}{\partial t} =  \nabla \cdot (\phi \mathbf{D} \nabla c) -  \nabla \cdot ( \frac{\mathbf{q} c}{S}) + Q_r, 
\end{equation}
\end{linenomath*}
where $c$ is the contaminant concentration [kg/m$^3$], $\mathbf{D}$ denotes the diffusion coefficient [m$^2$/s], and $Q_r$ is the source/sink term of salinity [kg/m$^3$/s]. 

\begin{figure}[!htb]
\centering
\includegraphics[trim = 0 0 0 0, clip, width=0.7\linewidth]{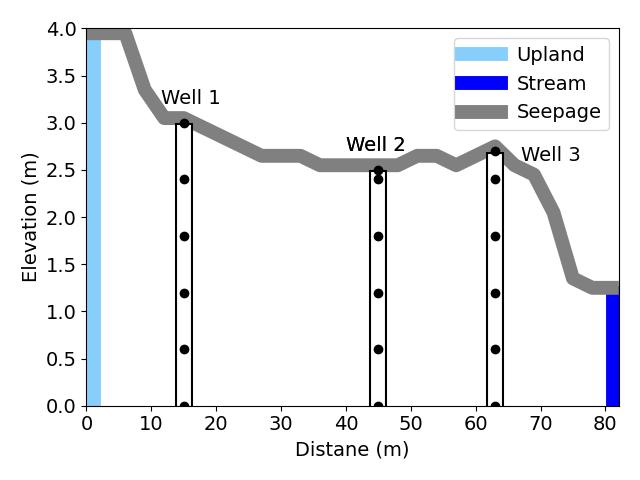}
\caption{Simulation setup for the Beaver Creek tidal stream-floodplain system. The model simulates subsurface flow and reactive transport using PFLOTRAN, driven by Dirichlet boundary conditions on the upland and stream, with seepage and no-flow boundary conditions applied to the ground surface and bottom, respectively. The initial simulation spans four years to reach steady-state groundwater conditions, followed by a 20-year seawater intrusion simulation. Three monitoring wells are used for validation throughout the simulation.}
\label{fig:Beaver_Creek}
\end{figure}

The system is a 2D cross-section from the upland to the stream with the same dimensions and elevation as the cross-section described in \citeA{yabusaki2020floodplain}. The flow model has dimensions of 84 m $\times$ 4 m and is defined on a grid with 28 $\times$ 40 cells. As shown in Fig.~\ref{fig:Beaver_Creek}, the flow system is driven by Dirichlet pressure boundary conditions on the upland and stream sides. In contrast, a seepage boundary condition is applied to the ground surface and a no-flow boundary condition at the bottom. The source of salinity is the tidal stream, with the boundary conditions set based on one-year observations from hillslope wells and stream data \cite{yabusaki2020floodplain}. For more details on the model, please refer to \citeA{yabusaki2020floodplain}.

\begin{figure}[!htb]
\centering
\begin{minipage}{.325\linewidth}\centering
\includegraphics[trim = 12 30 10 40, clip, width=\linewidth]{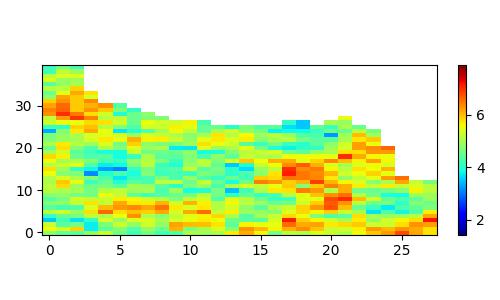}
\subcaption{Prior realization 1}
\end{minipage}
\begin{minipage}{.325\linewidth}\centering
\includegraphics[trim = 12 30 10 40, clip, width=\linewidth]{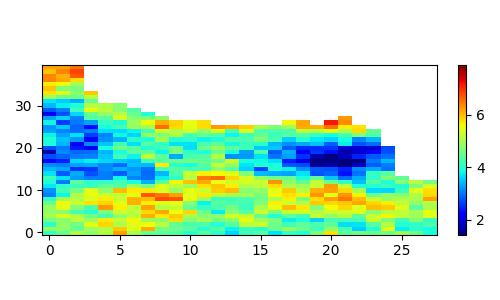}
\subcaption{Prior realization 2}
\end{minipage}
\begin{minipage}{.325\linewidth}\centering
\includegraphics[trim = 12 30 10 40, clip, width=\linewidth]{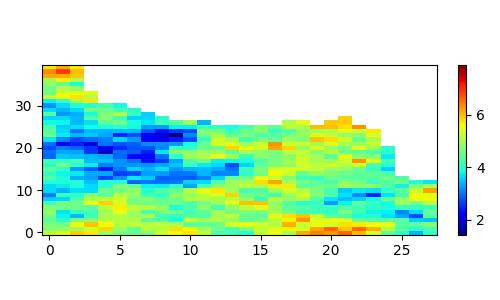}
\subcaption{Prior realization 3}
\end{minipage}

\caption{Log-permeability samples of three prior geomodels for PFLOTRAN simulation. Realization in (a) corresponds to the ``true" model used for data assimilation in Section \ref{sec:results}.}
\label{fig:prior_reals}
\end{figure}

Compared to the homogeneous case in \citeA{yabusaki2020floodplain}, we simulate a more complicated heterogeneous field in this work. Figure \ref{fig:prior_reals} present three random realizations of log-permeability. These Gaussian realizations are generated using sequential Gaussian simulation with SGeMS~\cite{remy2009applied}. The correlation lengths $l_x$ and $l_z$ are 5 and 5 grids, respectively. The mean and standard deviation of log-permeability (md) are 4.5 and 1, while the mean and standard deviation of porosity are 0.5 and 0.05. The same Gaussian field is used for both permeability and porosity, which serves as the input for the surrogate model. The other parameters are set based on \citeA{yabusaki2020floodplain}. The model in Fig. \ref{fig:prior_reals}(a) is used as the ``true" model for data assimilation in Section~\ref{sec:results}. 

The model is initially simulated for four years using annual observations to spin up the system to a dynamic steady state of groundwater levels, saturation, and total mass. After the spin-up, the simulation represents the initial freshwater conditions before seawater intrusion, which is then modeled over 7260 days ($\sim$20 years). Monitoring wells, as shown in Fig. \ref{fig:Beaver_Creek}, are placed in the same locations as in \cite{yabusaki2020floodplain} for model calibration and validation.

\subsection{Surrogate Model Development using U-FNO}
To efficiently address the uncertainty quantification and decision-making challenges in subsurface flow systems, it is necessary to discretize and solve equations~\ref{eq:gov_eq} to \ref{eq:advec_diff} thousands to millions of times, which can be computationally prohibitive. To overcome this challenge, we develop a deep-learning-based surrogate model, U-FNO, to approximate the flow and transport processes for various permeability and porosity fields under consistent boundary conditions. The flow simulation process is expressed as
\begin{linenomath*}
\begin{equation}\label{eq:simulation}
\mathbf{d} = [\mathbf{P}, \mathbf{c}] = f(\mathbf{m}),  
\end{equation}
\end{linenomath*}
where $\mathbf{m} \in \mathbb{R}^{n_c}$ denotes the geological model parameters (the permeability values in each cell, with $n_c = n_x \times n_z$ as the total number of cells, and $n_x$ and $n_z$ denoting the number of cells in the 2D x-z domain), $f$ is the flow and transport simulation process, and $\mathbf{d} $ represents the dynamic states of pressure $\mathbf{P}\in\mathbb{R}^{n_c \times n_t}$ and salinity $\mathbf{c}\in \mathbb{R}^{n_c \times n_t}$ in each cell over $n_t$ time steps. 

The forward process can be approximated by a surrogate model as
\begin{linenomath*}
\begin{equation}\label{eq:surrogate}
\mathbf{d} \approx \hat{\mathbf{d}} = [\hat{\mathbf{P}}, \hat{\mathbf{c}}] = \hat{f}(\mathbf{m}, \boldsymbol{\theta}),  
\end{equation}
\end{linenomath*}
where $\hat{\mathbf{d}} \in \mathbb{R}^{2n_c \times n_t}$ represents the approximate flow responses, $\hat{f}$ denotes the surrogate model, and $\boldsymbol{\theta}$ represents the parameters of the neural networks.

\begin{figure}[!htb]
\centering
\noindent\includegraphics[width=\textwidth]{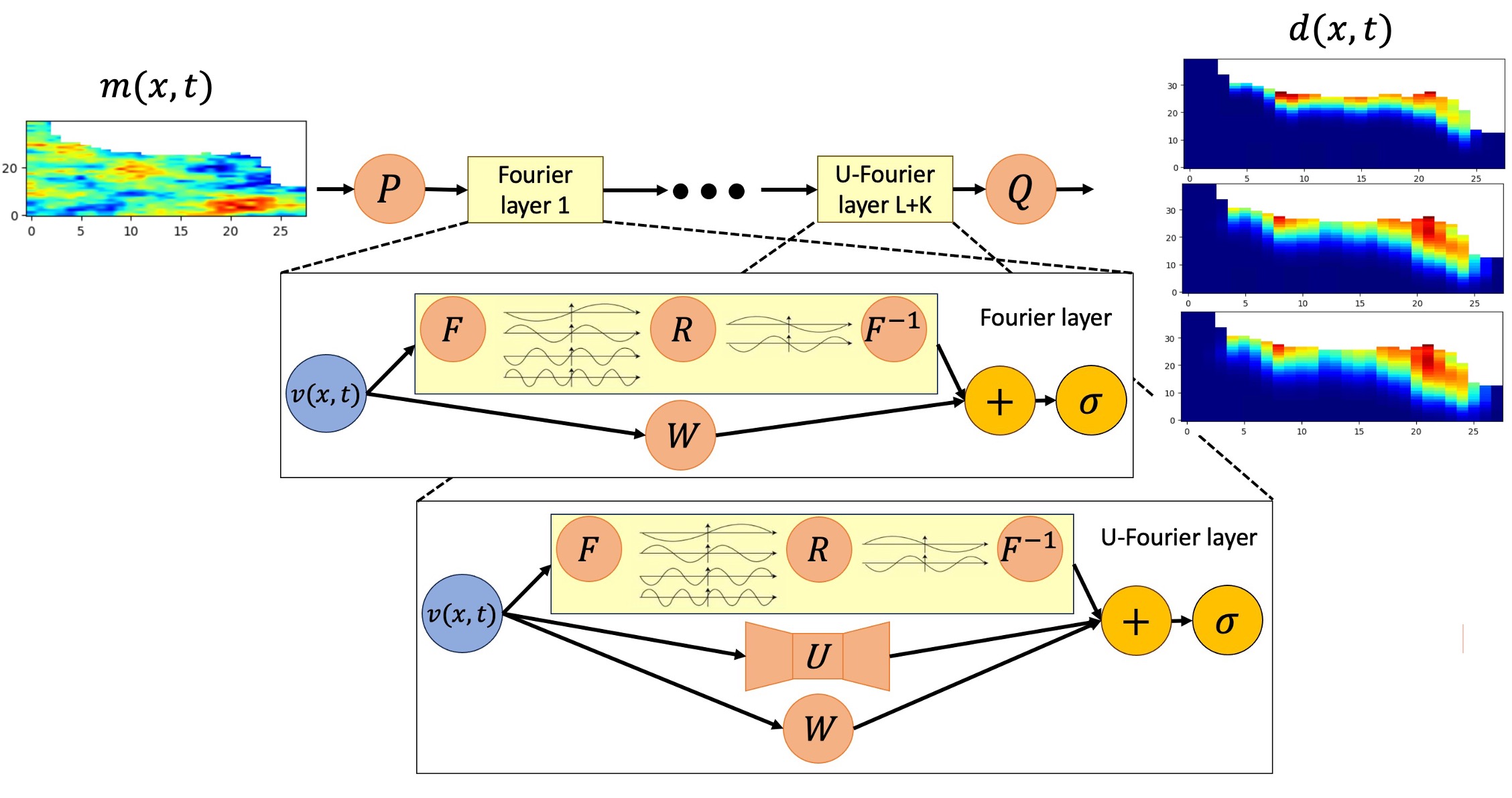}
\caption{Schematic of U-FNO, modified based on \protect\citeA{wen2022u}.}
\label{fig:u-fno}
\end{figure}

In this work, we apply the U-FNO framework proposed by~\citeA{wen2022u}. The U-FNO architecture is modified based on the FNO \cite{li2020fourier} with modifications that append a U-Net to the Fourier layer to improve prediction accuracy. U-FNO is designed to conduct an infinite-dimensional-space input output mapping. The architecture of U-FNO used in this work is presented in Fig.~\ref{fig:u-fno}. The geomodel input $\mathbf{m}$ is discretized on the 2D domain $\mathbf{x}$ and is repeated $n_t$ times as the function input $m(\mathbf{x}, t)$ for the networks. The input $a$ is then lifted to a higher-dimensional output $v_{0}(\mathbf{x}, t)$ through fully connected neural networks $P(\cdot)$. Multiple iterative Fourier layers and U-Fourier layers are then applied to capture the spatio-temporal features of flow states. Another fully connected layer $Q(\cdot)$ is applied in the end to map the U-Fourier layer output to surrogate model output $d(\mathbf{x}, t)$. 

In each Fourier layer $i$, $i = 1, \ldots, L$, the Fast Fourier Transform (FFT) is applied to the previous layer output $v_{i-1}(\mathbf{x}, t)$ to generate Fourier domain $\mathcal{F}(v_{i-1})$ and capture the latent features. A linear transformation $\mathcal{R}(\cdot)$ is then applied, followed by the inverse Fourier transform $\mathcal{F}^{-1}(\cdot)$ to map the Fourier domain back to the spatio-temporal domain. Additionally, a linear transform $W(\cdot)$ is applied directly to the function space $v_{i-1}(\mathbf{x}, t)$ and added to the output. The overall transformation process is written as 
\begin{linenomath*}
\begin{equation}\label{eq:fno_layer}
v_i (\mathbf{x}, t) = \sigma (\mathcal{F}^{-1}(\mathcal{R} \cdot \mathcal{F}(v_{i-1}))(\mathbf{x}, t) + W(v_{i-1}(\mathbf{x}, t))),  
\end{equation}
\end{linenomath*}
where $\sigma$ denotes the activation function, $L$ is the number of Fourier layer used in U-FNO. K U-Fourier layers are applied after $L$ Fourier layers. U-Net $U(\cdot)$ is attached in each Fourier layer to improve the surrogate model performance. The output $v_i (\mathbf{x}, t)$ of U-Fourier layer is expressed as 
\begin{linenomath*}
\begin{equation}\label{eq:ufno_layer}
v_i (\mathbf{x}, t) = \sigma (\mathcal{F}^{-1}(\mathcal{R} \cdot \mathcal{F}(v_{i-1}))(\mathbf{x}, t) + U(v_{i-1}(\mathbf{x}, t)) +  W(v_{i-1}(\mathbf{x}, t))),  
\end{equation}
\end{linenomath*}
for $i = L+1, \cdots, L+K$. The U-FNO architecture is written as 
\begin{linenomath*}
\begin{equation}\label{eq:ufno_layer}
m(\mathbf{x}, t) \rightarrow v_0 = P(m(\mathbf{x}, t)) \rightarrow \ldots \rightarrow v_L(\mathbf{x}, t) \rightarrow v_{L+K}(\mathbf{x}, t) \rightarrow Q(v_{L+K}) \rightarrow d(\mathbf{x}, t). 
\end{equation}
\end{linenomath*}
Please refer to~\citeA{wen2022u} for more details on U-FNO. 

The training objective is to minimize the difference between the flow simulation output with the surrogate predictions from U-FNO. The minimization process of U-FNO parameters $\boldsymbol{\theta}$ is described as 
\begin{linenomath*}
\begin{equation}\label{eq:obj_fun}
\boldsymbol{\theta} = \operatorname*{argmin}_{\boldsymbol{\theta}} \frac{1}{n_{smp}} \frac{1}{n_{t}} \sum_{i=1}^{n_{smp}} \sum_{i=1}^{n_{t}} ||\mathbf{d}_i^t  - \hat{\mathbf{d}}_i^t ||_2^2 + \lambda ||\mathbf{d}_i^{active, t}  - \hat{\mathbf{d}}_i^{active, t} ||_2^2, 
\end{equation}
\end{linenomath*}
where $n_{smp}$ is the number of training samples, $\lambda$ is the extra weight applied to the data in the active cells $\mathbf{d}^{active}$. These active cells are determined based on the digital elevation model for the 2D vertical cross-section. The adaptive moment estimation (ADAM) in~\cite{kingma2014adam} is applied to conduct the minimization process. 

\subsection{GeoFUSE: Data Assimilation Framework Utilizing ESMDA with Parameterization}\label{sec:GeoFUSE}
To improve flow prediction accuracy, we apply data assimilation to calibrate the geological model, employing the GeoFUSE framework (\textbf{Geo}logical \textbf{F}orecasting \textbf{u}sing \textbf{S}urrogates and \textbf{E}SMDA). GeoFUSE integrates PCA for dimensionality reduction, the U-FNO surrogate model for high-fidelity flow prediction, and the ESMDA for iterative parameter calibration. This combination enables efficient model calibration, enhancing prediction accuracy while reducing uncertainty. A Bayesian framework is employed within the data assimilation process to generate posterior probability distributions.

In practical applications, monitoring data is collected from wells or geophysical observations. For synthetic cases like ours, the observed data is generated by adding random noise to a synthetic ``true" model (Fig. \ref{fig:prior_reals}(a)) as follows
\begin{linenomath*}
\begin{equation}\label{eq:obs}
\mathbf{d}_\text{obs} = f(\mathbf{m}) + \boldsymbol{\epsilon},    
\end{equation}
\end{linenomath*}
where $\boldsymbol{\epsilon}$ is the measurement error. When no structural bias is considered, the measurement error $\boldsymbol{\epsilon}$ is assumed to be sampled from Gaussian distribution $N(\mathbf{0}, C_D)$, where $C_D$ denotes the error covariance matrix. The posterior probability density function of model parameter $\mathbf{m}$ conditioned to the observation $\mathbf{d}_{\text{obs}}$ using Bayesian framework is given by 
\begin{linenomath*}
\begin{equation}\label{eq:pdf}
p(\mathbf{m}|\mathbf{d}_\text{obs} ) = \frac{p(\mathbf{m}) p(\mathbf{d}_\text{obs} | \mathbf{m})}{ p(\mathbf{d}_\text{obs})} \propto p(\mathbf{m}) p(\mathbf{d}_\text{obs} | \mathbf{m}),  
\end{equation}
\end{linenomath*}
where $p(\mathbf{m})$ is the prior distribution of model parameters, $p(\mathbf{d}_\text{obs} | \mathbf{m})$ is the likelihood function, and $p(\mathbf{d}_\text{obs})$ is a normalization factor that can be ignored. 

\begin{figure}[!htb]
\noindent\includegraphics[width=\textwidth]{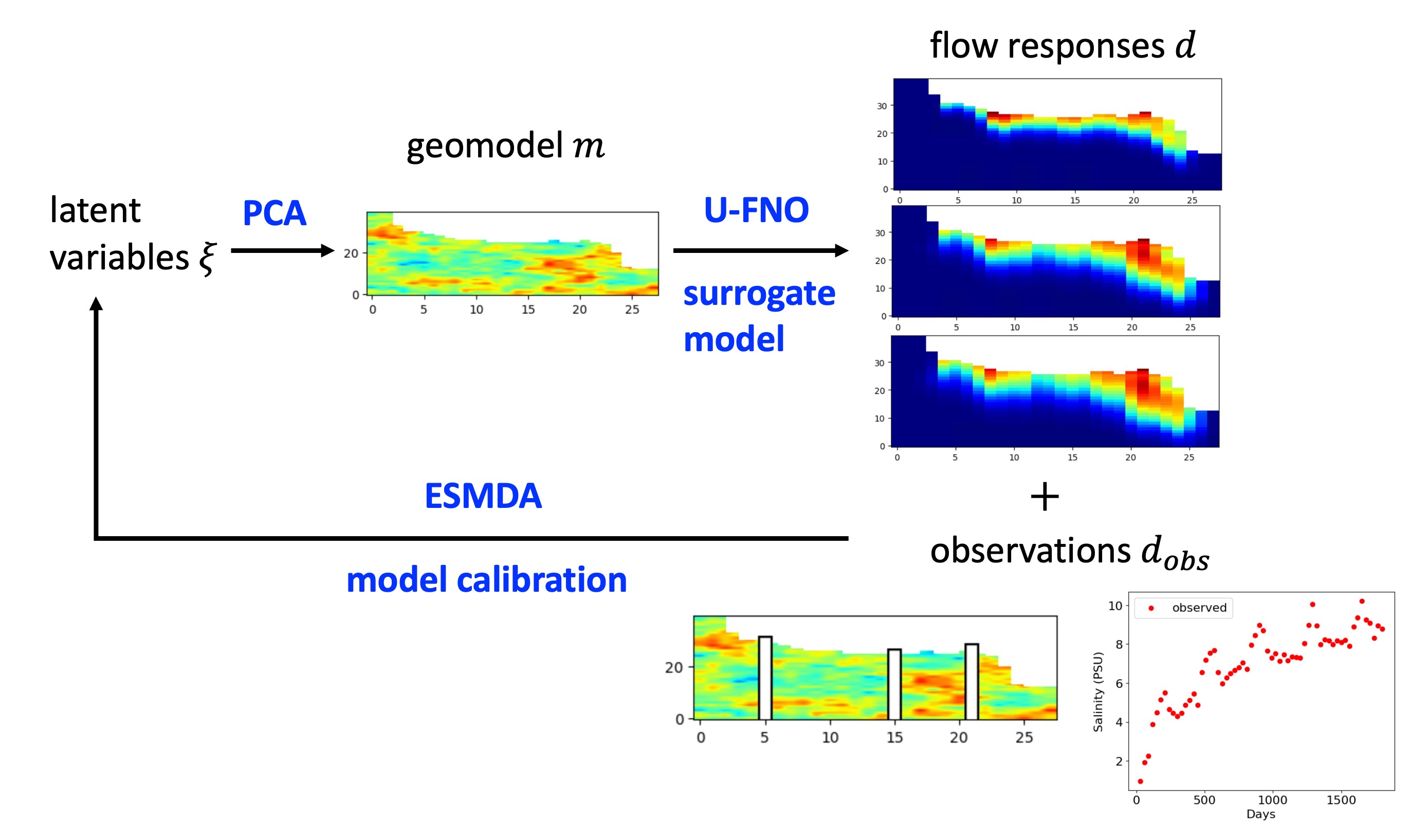}
\caption{Workflow of the GeoFUSE framewrok for data assimilation.}
\label{fig:pca-u-fno-esmda}
\end{figure}

Within the GeoFUSE framework as shown in Fig.~\ref{fig:pca-u-fno-esmda}, ESMDA \cite{emerick2013ensemble} is employed to sample posterior models, as illustrated in Eq.~\ref{eq:pdf}. Parameterization is generally needed to reduce the dimension of high-dimensional geological models and preserve the geological features. In this work, we consider a 2D Gaussian geomodel, and PCA is applied. The 2D geomodel can be generated from $\mathbf{m} \approx \mathbf{m}_\text{PCA}(\boldsymbol{\xi}) = \Phi \boldsymbol{\xi} +  \bar{\mathbf{m}}$, where $\boldsymbol {\xi} \in \mathbb{R}^{n_l}$ denotes the latent variables, $\Phi$ denotes the basis matrix generated from singular value decomposition, and $\bar{\mathbf{m}}$ is the average value of model parameters. Using U-FNO as a surrogate model, the flow responses can be generated from $\hat{f}(\mathbf{m}_\text{PCA}(\boldsymbol{\xi}))$. 

In the GeoFUSE data assimilation framework, the latent variables $\boldsymbol{\xi}$ are updated using 
\begin{linenomath*}
\begin{equation}\label{eq:esmda}
\boldsymbol{\xi}_i^{k+1} = \boldsymbol{\xi}_i^k + C_{\xi\mathbf{d}}^k(C^k_{\mathbf{d}} + \alpha_kC_D)^{-1}(\mathbf{d}_{\text{obs}} + \sqrt{\alpha_k}\mathbf{e}_i^k - \hat{f}(\mathbf{m}_\text{PCA}(\boldsymbol{\xi}))_i^k),  
\end{equation}
\end{linenomath*}
for $k = 1, \ldots, N_a$ and $i = 1, \ldots, N_r$, where $N_a$ is the number of data assimilation steps, and $N_r$ is the number of ensemble realizations. In this work, we use $N_a = 4$, and the inflation coefficients $\alpha_k$ are set to 4. And $C_{\xi\mathbf{d}}$ is the cross-covariance of latent variables $\boldsymbol{\xi}$ and data variables $\mathbf{d}$, and $C_{\mathbf{d}}$ is the data covariance matrix. These covariance matrices are updated at each assimilation step, corresponding to the updated ensemble. Random noise $\mathbf{e}$, sampled from a Gaussian distribution $N(\mathbf{0}, C_D)$, is added in each iteration for mismatch calculation. After all assimilation steps are completed, the posterior models $\mathbf{m}_\text{post}$ are generated using PCA on the posterior latent variables $\boldsymbol{\xi}_\text{post}$, and the posterior predictions $\mathbf{d}_\text{post}$ are generated using U-FNO on the posterior models.

Figure~\ref{fig:pca-u-fno-esmda} presents the GeoFUSE framework for data assimilation in the seawater intrusion problem. First, latent variables $\boldsymbol{\xi}$ are sampled from a Gaussian distribution $N(\mathbf{0}, I)$ to generate a prior ensemble with $N_r$ samples. PCA is then applied to the latent variables $\boldsymbol{\xi}$ to generate prior geomodels $\mathbf{m}$. The surrogate model is then applied to generate flow responses $\mathbf{d}$. ESMDA is performed, conditioned on the observations $\mathbf{d}_{\text{obs}}$, to calibrate the latent variables $\boldsymbol{\xi}$. In each iteration, PCA and U-FNO are applied to generate pressure and salinity data that match the observations. After $N_a$ iterations, we generate posterior models $\mathbf{m}_\text{post}$ and posterior predictions $\mathbf{d}_\text{post}$.

%% file: results.tex
\section{Results}\label{sec:results}
The following section presents the prediction results, focusing on the performance of the U-FNO surrogate model, comparisons with PFLOTRAN simulations, and the model calibration and uncertainty reduction achieved through the GeoFUSE framework. These results highlight improvements in computational efficiency and predictive accuracy.

\subsection{Surrogate Model (U-FNO) Training and Computational Efficiency}
We simulate 1600 prior realizations to collect pressure and salinity data at 243 time steps, with measurements taken every 30 days. Of these, 1500 realizations are used to train the U-FNO-based surrogate model, while the remaining 100 are reserved for evaluating its predictive performance. The U-FNO architecture, comprising over 126 million parameters, is employed to train separate models for pressure and salinity outputs (Table~\ref{tab:u-fno}).

The training process, conducted on a single Nvidia Tesla A100 GPU, takes 6.5 hours over 200 epochs for each network. In comparison, a full PFLOTRAN forward simulation required approximately 1.5 hours with one CPU. However, the U-FNO surrogate model predictions are computed in just 0.015 seconds, resulting in a computational speedup of around 360,000 times.

\begin{table}[!ht]
\footnotesize
\renewcommand{\arraystretch}{1} 
\centering
\caption{Architecture of U-FNO.} 
\begin{tabular}{l | l | l}
\hline
     Network & Layer & Output  \\
     \hline
     Input & & (40, 28, 243, 1) \\ 
     Padding & Padding &  (40, 32, 248, 36) \\
     Lifting & Linear & (40, 32, 248, 36) \\
     Fourier layer 1 & Fourier3d/Conv1d/Add/ReLu & (40, 32, 248, 36) \\
     Fourier layer 2 & Fourier3d/Conv1d/Add/ReLu & (40, 32, 248, 36) \\
     Fourier layer 3 & Fourier3d/Conv1d/Add/ReLu & (40, 32, 248, 36) \\
     U-Fourier layer 1 & Fourier3d/Conv1d/UNet3d/Add/ReLu & (40, 32, 248, 36) \\
     U-Fourier layer 2 & Fourier3d/Conv1d/UNet3d/Add/ReLu & (40, 32, 248, 36) \\
     U-Fourier layer 3 & Fourier3d/Conv1d/UNet3d/Add/ReLu & (40, 32, 248, 36) \\
     Projection 1 & Linear & (40, 32, 248, 128) \\
     Projection 2 & Linear & (40, 32, 248, 1) \\
     Depadding & & (40, 28, 243, 1)\\
     \hline
\end{tabular}
\label{tab:u-fno}
\end{table}

\subsection{Comparisons of U-FNO Predictions and PFLOTRAN Simulations}

\begin{figure}[!htb]
\centering
\begin{minipage}{.325\linewidth}\centering
\includegraphics[trim = 12 30 10 40, clip, width=\linewidth]{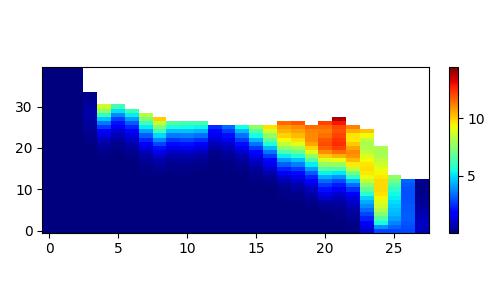}
\subcaption{Simulation (1050 days)}
\end{minipage}
\begin{minipage}{.325\linewidth}\centering
\includegraphics[trim = 12 30 10 40, clip, width=\linewidth]{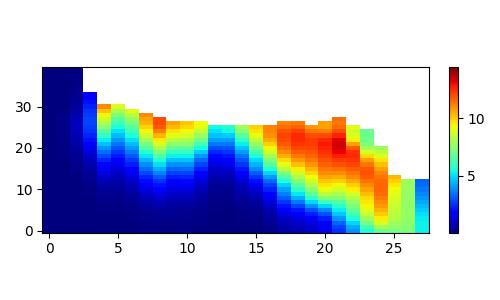}
\subcaption{Simulation (3570 days)}
\end{minipage}
\begin{minipage}{.325\linewidth}\centering
\includegraphics[trim = 12 30 10 40, clip, width=\linewidth]{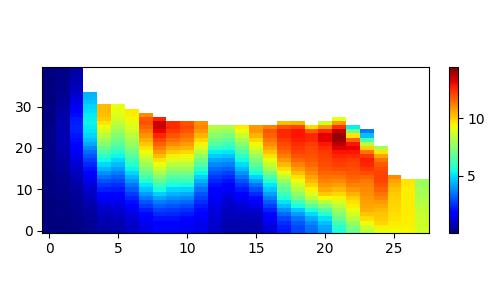}
\subcaption{Simulation (7170 days)}
\end{minipage}
\begin{minipage}{.325\linewidth}\centering
\includegraphics[trim = 12 30 10 40, clip, width=\linewidth]{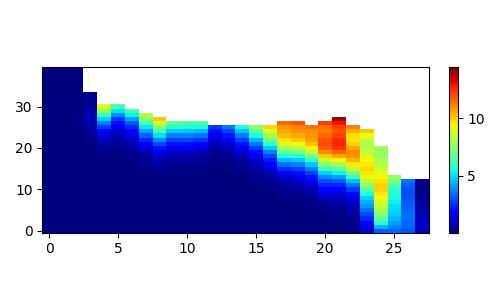}
\subcaption{Surrogate (1050 days)}
\end{minipage}
\begin{minipage}{.325\linewidth}\centering
\includegraphics[trim = 12 30 10 40, clip, width=\linewidth]{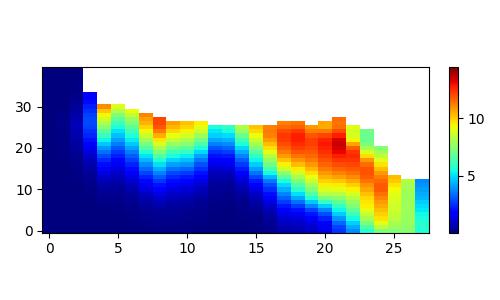}
\subcaption{Surrogate (3570 days)}
\end{minipage}
\begin{minipage}{.325\linewidth}\centering
\includegraphics[trim = 12 30 10 40, clip, width=\linewidth]{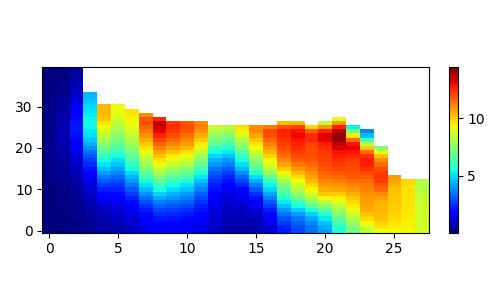}
\subcaption{Surrogate (7170 days)}
\end{minipage}
\begin{minipage}{.325\linewidth}\centering
\includegraphics[trim = 12 30 10 40, clip, width=\linewidth]{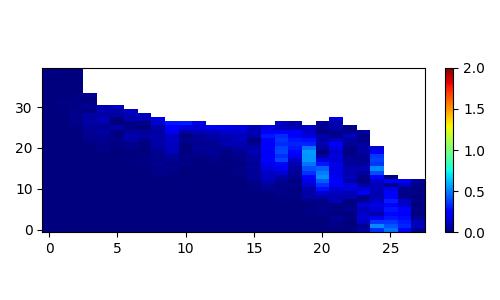}
\subcaption{Error (1050 days)}
\end{minipage}
\begin{minipage}{.325\linewidth}\centering
\includegraphics[trim = 12 30 10 40, clip, width=\linewidth]{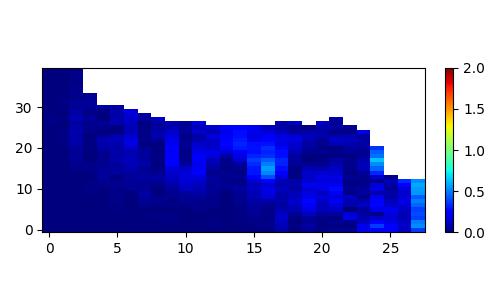}
\subcaption{Error (3570 days)}
\end{minipage}
\begin{minipage}{.325\linewidth}\centering
\includegraphics[trim = 12 30 10 40, clip, width=\linewidth]{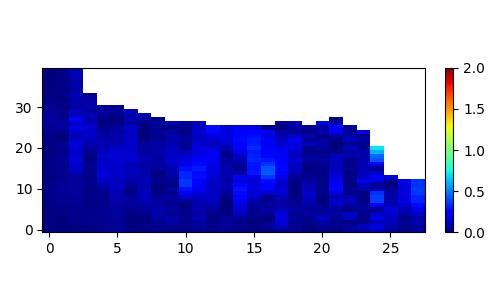}
\subcaption{Error (7170 days)}
\end{minipage}
\caption{Salinity along the cross-section for flow simulation (top row), the surrogate model (middle row), and the absolute error of predictions (bottom row) at 1050 days (early stage), 3570 days (mid stage), and 7170 days (late stage) for the prior realization 1 in Fig.~\ref{fig:prior_reals}(a). }
\label{fig:prior_salinity_time}
\end{figure}

\begin{figure}[!htb]
\centering
\begin{minipage}{.325\linewidth}\centering
\includegraphics[trim = 12 30 10 40, clip, width=\linewidth]{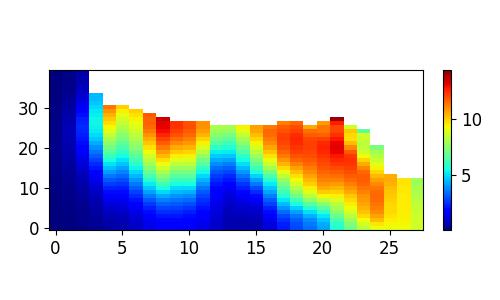}
\subcaption{Realization 1 (sim)}
\end{minipage}
\begin{minipage}{.325\linewidth}\centering
\includegraphics[trim = 12 30 10 40, clip, width=\linewidth]{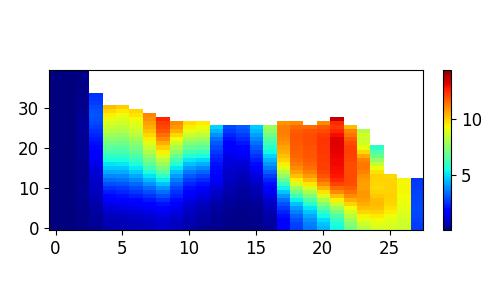}
\subcaption{Realization 2 (sim)}
\end{minipage}
\begin{minipage}{.325\linewidth}\centering
\includegraphics[trim = 12 30 10 40, clip, width=\linewidth]{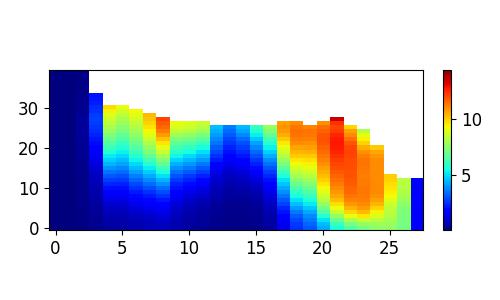}
\subcaption{Realization 3 (sim)}
\end{minipage}

\begin{minipage}{.325\linewidth}\centering
\includegraphics[trim = 12 30 10 40, clip, width=\linewidth]{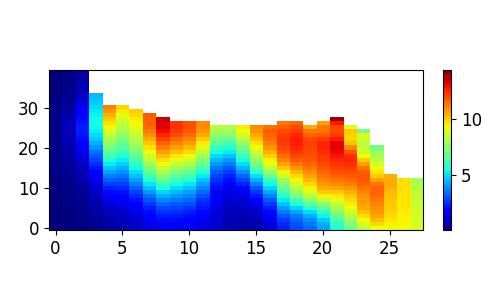}
\subcaption{Realization 1 (surr)}
\end{minipage}
\begin{minipage}{.325\linewidth}\centering
\includegraphics[trim = 12 30 10 40, clip, width=\linewidth]{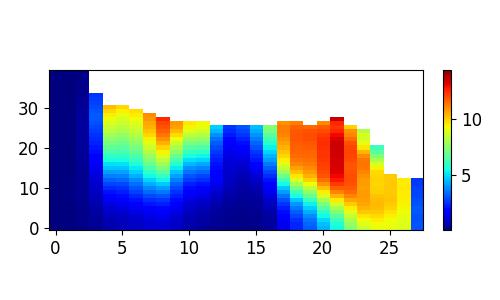}
\subcaption{Realization 2 (surr)}
\end{minipage}
\begin{minipage}{.325\linewidth}\centering
\includegraphics[trim = 12 30 10 40, clip, width=\linewidth]{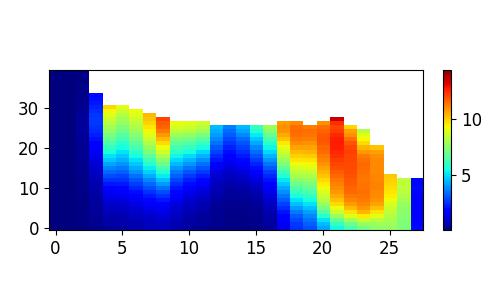}
\subcaption{Realization 3 (surr)}
\end{minipage}

\begin{minipage}{.325\linewidth}\centering
\includegraphics[trim = 12 30 10 40, clip, width=\linewidth]{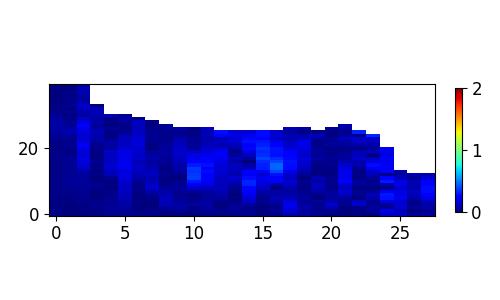}
\subcaption{Realization 1 (error)}
\end{minipage}
\begin{minipage}{.325\linewidth}\centering
\includegraphics[trim = 12 30 10 40, clip, width=\linewidth]{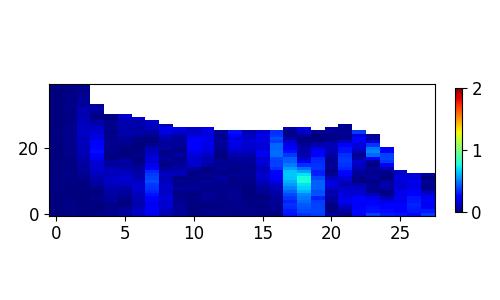}
\subcaption{Realization 2 (error)}
\end{minipage}
\begin{minipage}{.325\linewidth}\centering
\includegraphics[trim = 12 30 10 40, clip, width=\linewidth]{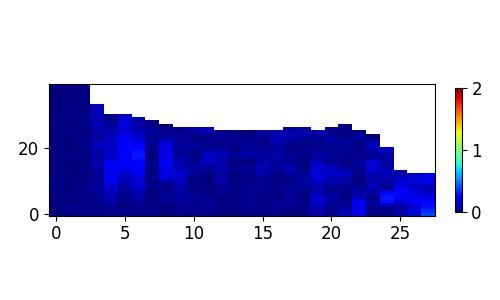}
\subcaption{Realization 3 (error)}
\end{minipage}

\caption{Salinity along the cross-section for flow simulation (top row), the surrogate model (middle row), and the absolute error of predictions (bottom row) for the three prior realizations in Fig.~\ref{fig:prior_reals}. Results are for flow results at 7260 days (late stage).}
\label{fig:prior_salinity_multiple} 
\end{figure}

To assess the predictive accuracy of the U-FNO surrogate model, we compare its results with the PFLOTRAN simulations. Figures \ref{fig:prior_salinity_time} and \ref{fig:prior_salinity_multiple} demonstrate the close agreement between the predicted and simulated salinity distributions over time and across multiple realizations in Fig.~\ref{fig:prior_reals}. The top row presents PFLOTRAN results, the middle row shows U-FNO predictions, and the bottom row highlights the absolute errors between the two. The comparisons are made for three key time points: 1050 days, representing the early-stage (approximately 3 years), 3570 days, marking the mid-stage (approximately 10 years), and 7170 days, indicating the late-stage (approximately 20 years), clearly illustrating the salinity dynamics over the 20-year period. As shown in Fig. \ref{fig:prior_salinity_time}, the U-FNO model effectively captures both the temporal and spatial dynamics of salinity. The absolute error for this realization falls in the middle range of the 100 test samples, demonstrating the model's robustness in approximating complex flow and transport processes.

Figure \ref{fig:prior_salinity_multiple} presents the variability in salinity plume size and shape across three different permeability realizations. This variability highlights the challenges of modeling heterogeneous systems, yet U-FNO successfully replicates these dynamics. The surrogate model consistently delivers high accuracy in its predictions, with the errors between U-FNO and PFLOTRAN being notably small, as shown in the bottom rows showing errors.

\begin{figure}[!htb]
\centering
\noindent\includegraphics[width=0.65\textwidth]{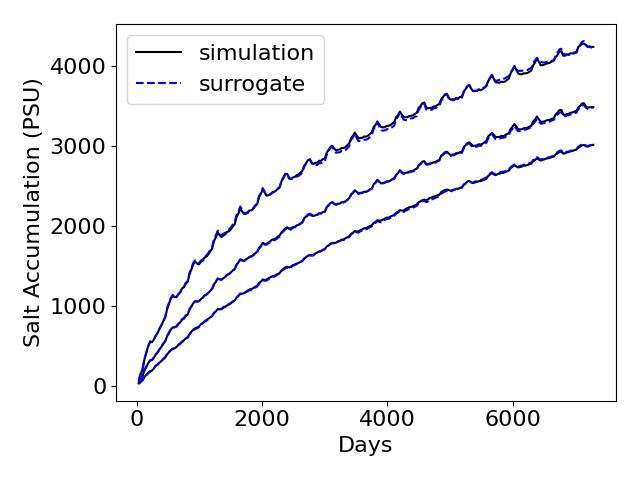}
\caption{Comparison of salinity accumulation from simulation results (black curves) and from surrogate results (blue dashed curves). Lower, middle and upper curves are P$_{10}$, P$_{50}$, and P$_{90}$ results}
\label{fig:prior_stats}
\end{figure}

In addition to salinity distribution, Fig. ~\ref{fig:prior_stats} compares salinity accumulation between the U-FNO surrogate predictions and PFLOTRAN simulations for 100 test realizations. The results show the 10\textsuperscript{th}, 50\textsuperscript{th}, and 90\textsuperscript{th} percentiles (P$_{10}$, P$_{50}$, and P$_{90}$) over a period of 7,260 days. The black curves represent the PFLOTRAN results, while the blue dashed curves represent the U-FNO predictions. Over the 20-year period, the consistent increase in salinity accumulation and the close agreement between the surrogate model and full-scale simulations demonstrate the accuracy of U-FNO in approximating these long-term cumulative effects.

\section{Uncertainty Reduction Through GeoFuse: Data Assimilation and Model Calibration}
In the data assimilation process, pressure and salinity data are collected from three monitors at the locations shown as the black dots (18 monitoring locations in total) in Fig.~\ref{fig:Beaver_Creek}. Data are collected monthly for the first five years, resulting in 2160 observations. To simulate realistic environmental conditions, Gaussian noise is added to the flow simulation results. We add Gaussian noise to the flow simulation results for the ``true" model in Fig.~\ref{fig:prior_reals}(a). The standard deviation for the water level is set to 0.005 m, and for salinity, it is set to 0.2. 

GeoFUSE effectively handles the high-dimensional geomodel parameters (permeability in each grid block) by reducing their dimension and enabling efficient parameter updates, with the latent dimension after PCA set to 642 based on energy criteria. The GeoFUSE framework applies ESMDA to assimilate observations over $N_a = 4$ steps and calibrate the geological models. The data assimilation process results in more accurate predictions throughout the 20-year simulation period, conditioned on measurements from monitoring wells. The GeoFUSE framework significantly reduces the computational costs while delivering improved model and prediction accuracy for managing complicated subsurface environments.

\begin{figure}[!htb]
\centering
\begin{minipage}{.47\linewidth}\centering
\includegraphics[trim = 12 70 5 70, clip, width=\linewidth]{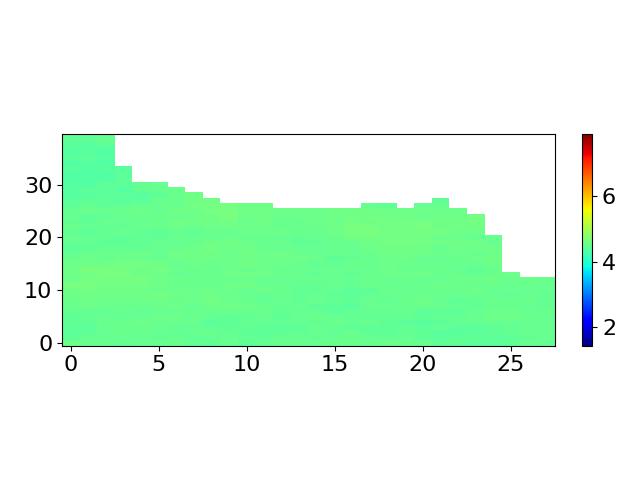}
\subcaption{Mean of prior models}
\end{minipage}
\begin{minipage}{.47\linewidth}\centering
\includegraphics[trim = 12 70 5 70, clip, width=\linewidth]{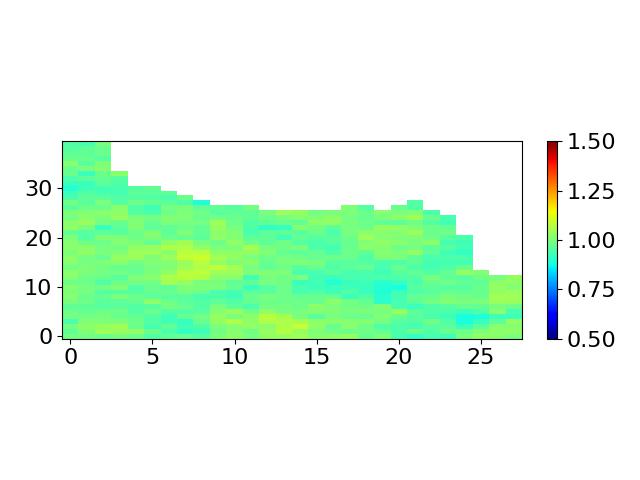}
\subcaption{Std of prior models}
\end{minipage}
\centering
\begin{minipage}{.47\linewidth}\centering
\includegraphics[trim = 12 70 5 70, clip, width=\linewidth]{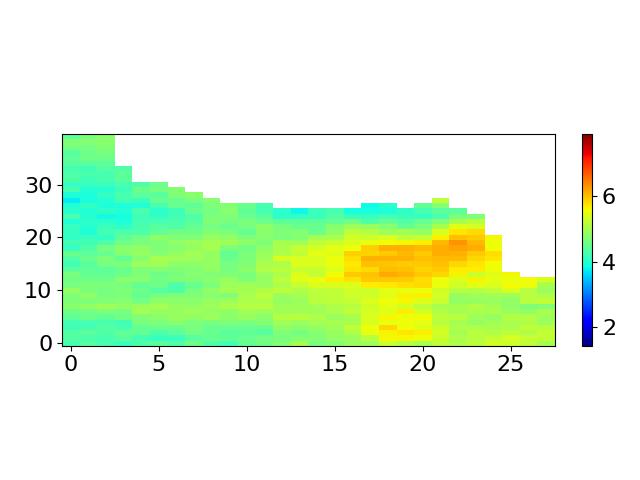}
\subcaption{Mean of posterior models}
\end{minipage}
\begin{minipage}{.47\linewidth}\centering
\includegraphics[trim = 12 70 5 70, clip, width=\linewidth]{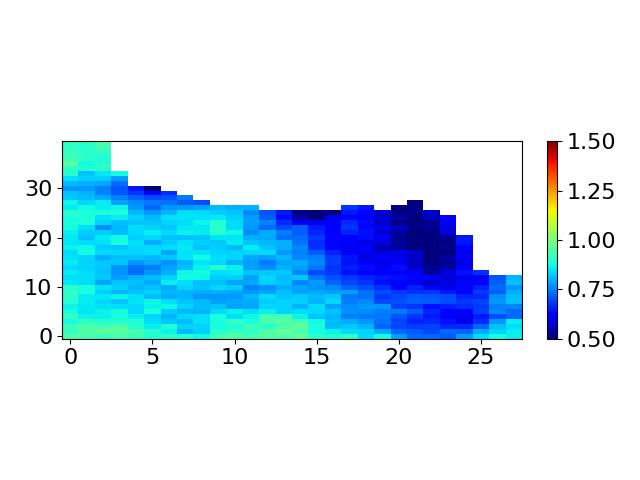}
\subcaption{Std of posterior models}
\end{minipage}
\caption{Prior and posterior mean and standard deviation for geological models.}
\label{fig:post_geomodel}
\end{figure}

Figure \ref{fig:post_geomodel} shows the prior and posterior means and standard deviations of log-permeability. The prior models follow a Gaussian distribution, with the prior means close to the mean value of 4.5 and standard deviations near 1. The posterior means of the log-permeability parameters calibrated by GeoFUSE are closely aligned with the true geological model. The posterior models demonstrate substantial uncertainty reduction compared to the prior models, particularly in regions near the stream with high permeability. Additionally, the standard deviations of the posterior models are significantly reduced in areas of high permeability, indicating a marked decrease in uncertainty. These improvements highlight the effectiveness of the GeoFUSE framework in calibrating the geological model.

\begin{figure}[!htb]
\centering
\centering
\begin{minipage}{.47\linewidth}\centering
\includegraphics[trim = 0 0 0 0, clip, width=\linewidth]{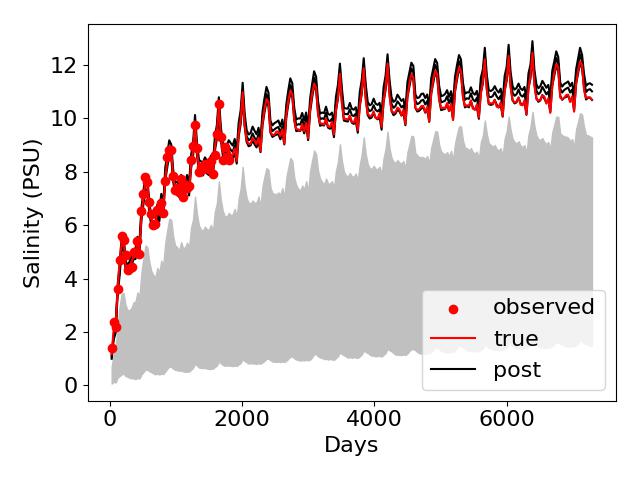}
\subcaption{Salinity at well 2}
\end{minipage}
\begin{minipage}{.47\linewidth}\centering
\includegraphics[trim = 0 0 0 0, clip, width=\linewidth]{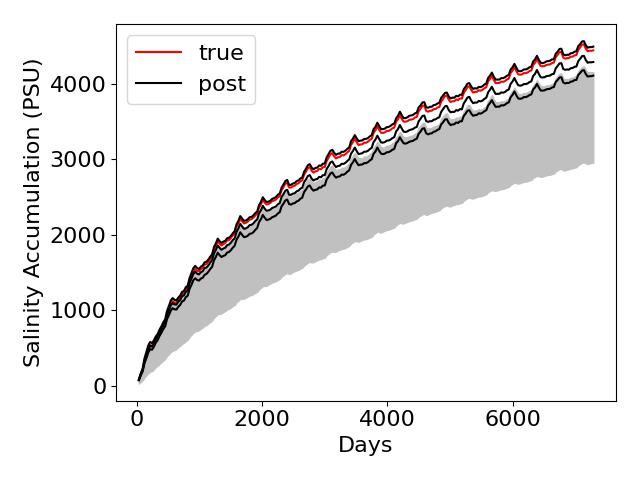}
\subcaption{Salinity accumulation}
\end{minipage}
\caption{Prior and posterior statistics for salinity at well 2, and salinity accumulation for ``true" model. Lower, middle and upper curves are P$_{10}$, P$_{50}$, and P$_{90}$ results}
\label{fig:post_flow_stats}
\end{figure}

Results for salinity at the land surface of well 2 and salinity accumulation are shown in Fig.~\ref{fig:post_flow_stats}. Salinity accumulation refers to the total salinity in the entire region. The gray shaded area shows the P$_{10}$ - P$_{90}$ range of the prior. The black curves show the P$_{10}$, P$_{50}$, and P$_{90}$ posterior results. The red dots show the observations, including random error, and the red curves show the true flow responses. Significant uncertainty reduction is achieved for both salinity at the well and salinity accumulation. Notably, even when the true salinity values fall outside the prior range, the posterior results still show substantial uncertainty reduction. These findings demonstrate the effectiveness of GeoFUSE in improving the accuracy and reliability of salinity predictions over the 20-year simulation period.

\begin{figure}[!htb]
\centering
\begin{minipage}{.47\linewidth}\centering
\includegraphics[trim = 12 45 10 45, clip, width=\linewidth]{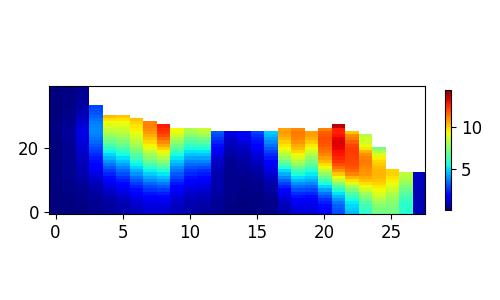}
\subcaption{Realization~1}
\end{minipage}
\begin{minipage}{.47\linewidth}\centering
\includegraphics[trim = 12 45 10 45, clip, width=\linewidth]{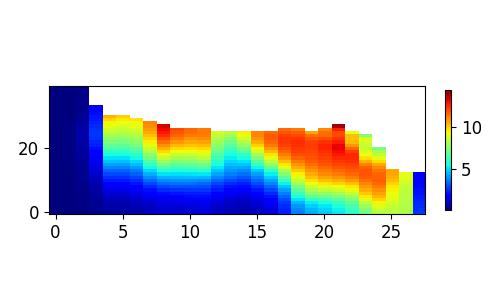}
\subcaption{Realization~1}
\end{minipage}

\begin{minipage}{.47\linewidth}\centering
\includegraphics[trim = 12 45 10 45, clip, width=\linewidth]{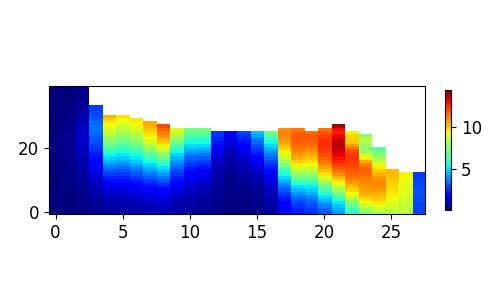}
\subcaption{Realization~2}
\end{minipage}
\begin{minipage}{.47\linewidth}\centering
\includegraphics[trim = 12 45 10 45, clip, width=\linewidth]{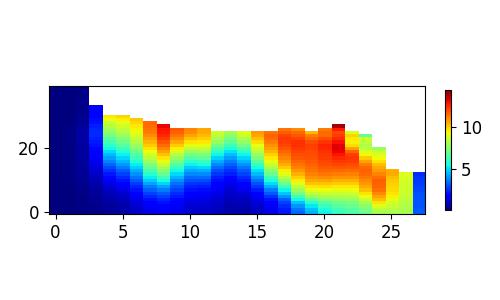}
\subcaption{Realization~2}
\end{minipage}

\begin{minipage}{.47\linewidth}\centering
\includegraphics[trim = 12 45 10 45, clip, width=\linewidth]{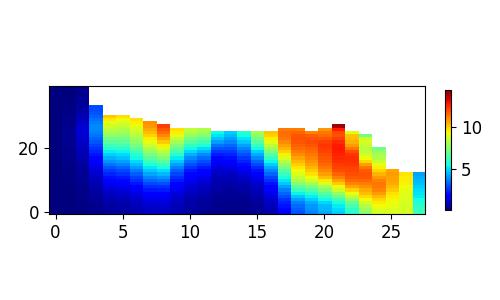}
\subcaption{Realization~3}
\end{minipage}
\begin{minipage}{.47\linewidth}\centering
\includegraphics[trim = 12 45 10 45, clip, width=\linewidth]{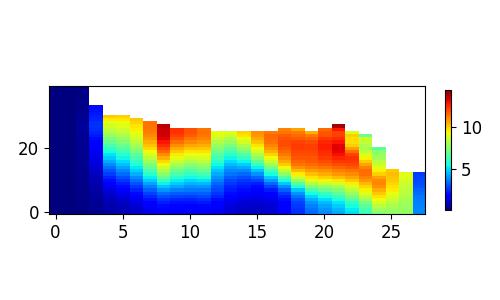}
\subcaption{Realization~3}
\end{minipage}

\begin{minipage}{.47\linewidth}\centering
\includegraphics[trim = 12 45 10 45, clip, width=\linewidth]{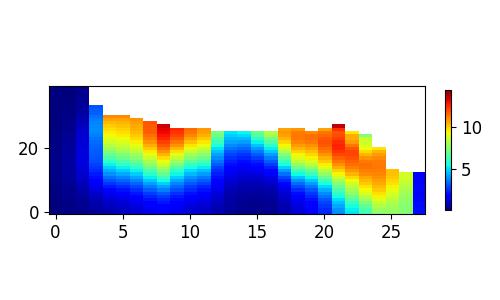}
\subcaption{Realization~4}
\end{minipage}
\begin{minipage}{.47\linewidth}\centering
\includegraphics[trim = 12 45 10 45, clip, width=\linewidth]{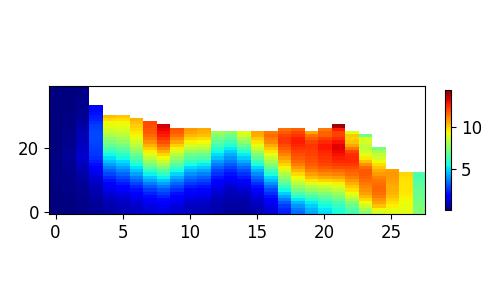}
\subcaption{Realization~4}
\end{minipage}

\begin{minipage}{.47\linewidth}\centering
\includegraphics[trim = 12 45 10 45, clip, width=\linewidth]{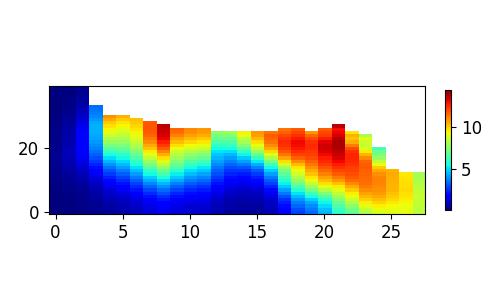}
\subcaption{Realization~5}
\end{minipage}
\begin{minipage}{.47\linewidth}\centering
\includegraphics[trim = 12 45 10 45, clip, width=\linewidth]{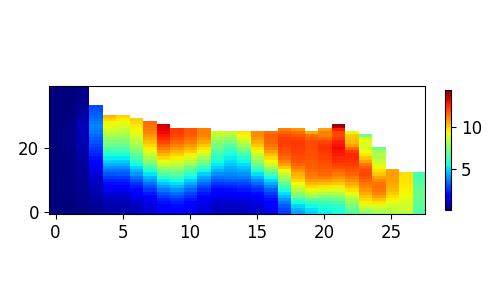}
\subcaption{Realization~5}
\end{minipage}
\caption{Representative prior (left column) and posterior (right column) salinity fields for true model~1 at 7260 days. True salinity field for this case shown in Fig.~\ref{fig:prior_salinity_multiple}(a).} \label{fig:post_s_map}
\end{figure}

The clustering analysis using the K-means and K-medoids methods allows for generating representative salinity fields from both prior and posterior models. The K-means method is first applied to generate five clusters, and K-medoids is then used to select the center for each cluster. Figure~\ref{fig:post_s_map} depicts the salinity fields from five representative clusters, showing how the posterior models more accurately capture the size and shape of the salinity plumes compared to the prior models. After applying GeoFUSE, the posterior models closely resemble the true salinity distributions, particularly in capturing larger and more distinct salinity plumes. This further demonstrates GeoFUSE’s capability to reduce uncertainty in the spatial distributions of state variables and effectively improve predictive accuracy.

These results collectively highlight the success of the GeoFUSE framework in significantly reducing computational costs, calibrating and reducing the uncertainty of geological models, and providing more reliable predictions of salinity distribution over the 20-year simulation period.

%% file: discussion.tex
\section{Discussion}
\subsection{Surrogate Model Accuracy and Limitations}
The development of the U-FNO model represents a significant advancement in computational efficiency and accuracy for subsurface flow simulations. It achieves a remarkable speedup of approximately 360,000 times over traditional simulation models like PFLOTRAN, offering substantial benefits for large-scale simulations where computational resources and time are limited \cite{wen2022u}. This efficiency is especially critical for scenarios requiring extensive Monte Carlo sampling for uncertainty quantification. As noted by \citeA{zhong2019predicting}, their cDC-GAN model similarly reduced simulation time, enabling numerous simulations for uncertainty analysis. However, U-FNO faces challenges in accurately predicting saltwater front movements, where its accuracy diminishes. This issue is not unique to U-FNO, as convolutional neural networks (CNN)-based methods \cite{mo2019deep, tang2020deep} have exhibited similar discrepancies when approximating multiphase flow behaviors in heterogeneous systems. Despite these challenges, U-FNO’s overall predictive accuracy remains impressive, making it a transformative tool for large-scale, data-intensive studies.

\subsection{Effectiveness of GeoFUSE}
The GeoFUSE framework substantially accelerates the subsurface flow modeling and data assimilation process while significantly reducing the uncertainty in geological models and long-term predictions, particularly for seawater intrusion and groundwater management in coastal areas. This approach aligns with previous studies that emphasize the importance of addressing geological uncertainty to enhance predictive capabilities \cite{sonnenborg2015climate, he2015assessing}. Within the GeoFUSE framework, PCA is used to reduce the dimension of geological models while preserving essential features, enabling more efficient uncertainty quantification for large-scale Gaussian models. Additionally, the ability of ESMDA to assimilate sparse and noisy data makes the framework both simple and robust for subsurface flow predictions. The results, which show improved salinity distribution predictions over a 20-year simulation period, further highlight the framework's effectiveness in model calibration and uncertainty reduction. Overall, GeoFUSE provides a transformative approach to subsurface flow simulations and uncertainty quantification in complex coastal environments by accelerating flow and transport simulations, efficiently calibrating geological models, and reducing the uncertainty of future predictions.

\subsection{Implications for Seawater Intrusion and Floodplain Management}
GeoFUSE offers a powerful tool for decision-makers involved in floodplain management, tidal restoration, and salinity control by providing rapid and accurate predictions and model calibrations. This capability enhances water resource management in the face of rising sea levels and climate change. By offering predictive modeling for salinity intrusion in vulnerable coastal regions, GeoFUSE plays a critical role in broader coastal management efforts. Its ability to deliver timely, accurate insights makes it essential for adaptive management, especially as seawater intrusion intensifies due to climate change. This allows decision-makers to respond swiftly to evolving conditions in coastal aquifers.

The integration of monitoring data into predictive models, as demonstrated in this study, significantly reduces model uncertainty and improves the management of coastal aquifers. GeoFUSE’s use of a surrogate model reduces computational costs, enabling rapid predictions for uncertainty quantification and decision-making. By assimilating monitoring data, the framework improves salinity forecasts by reducing uncertainty in the geological models. The model's performance, validated against PFLOTRAN simulation results and prior statistics, confirms its predictive accuracy and robustness in model calibration.

\subsection{Future Work, Model Scalability, and Caveats}
GeoFUSE has demonstrated substantial potential in subsurface flow modeling and uncertainty quantification, but there is room for further improvement and scalability. One area of focus is refining the U-FNO model, which has effectively captured complex multiphase flow dynamics \cite{wen2022u}. However, further adjustments could enhance its ability to account for fine-scale geological variations, which are crucial in heterogeneous environments. Additionally, incorporating more complex factors, such as boundary conditions influenced by climate change, could increase the model's robustness. Climate-induced variability significantly impacts hydrological cycles and groundwater dynamics \cite{he2015assessing}, and addressing these complexities would enable GeoFUSE to more accurately simulate the intricate interactions between geological formations and hydrological processes across diverse environments.

Future research could also focus on extending GeoFUSE to three-dimensional (3D) models, which are essential for accurately simulating subsurface flow in coastal regions increasingly affected by seawater intrusion. Recent studies highlight the importance of 3D geological modeling in understanding fluid dynamics and permeability in geothermal systems \cite{siler2019three, torresan20203d}. A 3D framework would better integrate geological, geophysical, and geochemical data, providing a more comprehensive approach to evaluating both anthropogenic and natural influences on subsurface flow \cite{strati2017perceiving, dwivedi2018hot}. By improving scalability and addressing these key areas, GeoFUSE could evolve in a versatile tool for groundwater management in diverse environments, particularly in coastal regions facing challenges such as groundwater depletion and climate change \cite{messier2015estimation, ebong2020stochastic}.

%% file: conclusion.tex
\section{Concluding Remarks}\label{sec:conclusion}
In this work, we developed a novel framework, GeoFUSE, to address the computational challenges associated with seawater intrusion modeling in coastal aquifers. GeoFUSE integrates the U-FNO surrogate model for fast spatio-temporal predictions with ESMDA for data assimilation and model calibration. The framework was applied to a 2D cross-section of a tidal stream-floodplain system, predicting pressure and salinity responses while accounting for geological heterogeneity and uncertainty in flow responses. The GeoFUSE approach enabled efficient data assimilation, with PCA parameterizing Gaussian geological models and ESMDA calibrating heterogeneous models for posterior sampling.

We trained the U-FNO model using 1500 realizations and demonstrated its ability to significantly reduce computational costs compared to PFLOTRAN simulations while maintaining predictive accuracy. The data assimilation framework was used to calibrate the geological models based on five years of observation data from three wells and predict salinity distribution and accumulation over a 20-year period. Results showed substantial uncertainty reduction for both models and flow responses, highlighting the effectiveness and efficiency of GeoFUSE.

Future work will focus on extending GeoFUSE to large 3D models and incorporating more complex boundary conditions, including the impacts of climate change, sea-level rise, and extreme weather events. Expanding the framework to include pumping and injection strategies for seawater intrusion mitigation, as well as uncertainty quantification for optimization, will further enhance real-time predictions and decision-making processes. While training deep learning models like U-FNO can introduce uncertainty due to the variability in training data from geological models, GeoFUSE holds great promise for advancing large-scale subsurface flow and transport simulations with improved accuracy.

%% file: main.bbl
\begin{thebibliography}{}

\bibitem [\protect \citeauthoryear {%
Arora%
, Mohanty%
\BCBL {}\ \BBA {} McGuire%
}{%
Arora%
\ \protect \BOthers {.}}{%
{\protect \APACyear {2012}}%
}]{%
arora2012uncertainty}
\APACinsertmetastar {%
arora2012uncertainty}%
\begin{APACrefauthors}%
Arora, B.%
, Mohanty, B.%
\BCBL {}\ \BBA {} McGuire, J.%
\end{APACrefauthors}%
\unskip\
\newblock
\APACrefYearMonthDay{2012}{}{}.
\newblock
{\BBOQ}\APACrefatitle {Uncertainty in dual permeability model parameters for structured soils} {Uncertainty in dual permeability model parameters for structured soils}.{\BBCQ}
\newblock
\APACjournalVolNumPages{Water Resources Research}{48}{1}{}.
\PrintBackRefs{\CurrentBib}

\bibitem [\protect \citeauthoryear {%
Arora%
, Mohanty%
\BCBL {}\ \BBA {} McGuire%
}{%
Arora%
\ \protect \BOthers {.}}{%
{\protect \APACyear {2011}}%
}]{%
arora2011inverse}
\APACinsertmetastar {%
arora2011inverse}%
\begin{APACrefauthors}%
Arora, B.%
, Mohanty, B\BPBI P.%
\BCBL {}\ \BBA {} McGuire, J\BPBI T.%
\end{APACrefauthors}%
\unskip\
\newblock
\APACrefYearMonthDay{2011}{}{}.
\newblock
{\BBOQ}\APACrefatitle {Inverse estimation of parameters for multidomain flow models in soil columns with different macropore densities} {Inverse estimation of parameters for multidomain flow models in soil columns with different macropore densities}.{\BBCQ}
\newblock
\APACjournalVolNumPages{Water resources research}{47}{4}{}.
\PrintBackRefs{\CurrentBib}

\bibitem [\protect \citeauthoryear {%
Bhattacharjya%
\ \BBA {} Datta%
}{%
Bhattacharjya%
\ \BBA {} Datta%
}{%
{\protect \APACyear {2009}}%
}]{%
bhattacharjya2009ann}
\APACinsertmetastar {%
bhattacharjya2009ann}%
\begin{APACrefauthors}%
Bhattacharjya, R\BPBI K.%
\BCBT {}\ \BBA {} Datta, B.%
\end{APACrefauthors}%
\unskip\
\newblock
\APACrefYearMonthDay{2009}{}{}.
\newblock
{\BBOQ}\APACrefatitle {{ANN-GA-based} model for multiple objective management of coastal aquifers} {{ANN-GA-based} model for multiple objective management of coastal aquifers}.{\BBCQ}
\newblock
\APACjournalVolNumPages{Journal of Water Resources Planning and Management}{135}{5}{314--322}.
\PrintBackRefs{\CurrentBib}

\bibitem [\protect \citeauthoryear {%
Cao%
, Zhang%
, Gan%
, Nan%
\BCBL {}\ \BBA {} Lu%
}{%
Cao%
\ \protect \BOthers {.}}{%
{\protect \APACyear {2024}}%
}]{%
cao2024deep}
\APACinsertmetastar {%
cao2024deep}%
\begin{APACrefauthors}%
Cao, C.%
, Zhang, J.%
, Gan, W.%
, Nan, T.%
\BCBL {}\ \BBA {} Lu, C.%
\end{APACrefauthors}%
\unskip\
\newblock
\APACrefYearMonthDay{2024}{}{}.
\newblock
{\BBOQ}\APACrefatitle {A deep learning-based data assimilation approach to characterizing coastal aquifers amid non-linearity and non-{G}aussianity challenges} {A deep learning-based data assimilation approach to characterizing coastal aquifers amid non-linearity and non-{G}aussianity challenges}.{\BBCQ}
\newblock
\APACjournalVolNumPages{Water Resources Research}{60}{7}{e2023WR036899}.
\PrintBackRefs{\CurrentBib}

\bibitem [\protect \citeauthoryear {%
Dodangeh%
, Rajabi%
, Carrera%
\BCBL {}\ \BBA {} Fahs%
}{%
Dodangeh%
\ \protect \BOthers {.}}{%
{\protect \APACyear {2022}}%
}]{%
dodangeh2022joint}
\APACinsertmetastar {%
dodangeh2022joint}%
\begin{APACrefauthors}%
Dodangeh, A.%
, Rajabi, M\BPBI M.%
, Carrera, J.%
\BCBL {}\ \BBA {} Fahs, M.%
\end{APACrefauthors}%
\unskip\
\newblock
\APACrefYearMonthDay{2022}{}{}.
\newblock
{\BBOQ}\APACrefatitle {Joint identification of contaminant source characteristics and hydraulic conductivity in a tide-influenced coastal aquifer} {Joint identification of contaminant source characteristics and hydraulic conductivity in a tide-influenced coastal aquifer}.{\BBCQ}
\newblock
\APACjournalVolNumPages{Journal of Contaminant Hydrology}{247}{}{103980}.
\PrintBackRefs{\CurrentBib}

\bibitem [\protect \citeauthoryear {%
Dwivedi%
, Arora%
, Steefel%
, Dafflon%
\BCBL {}\ \BBA {} Versteeg%
}{%
Dwivedi%
\ \protect \BOthers {.}}{%
{\protect \APACyear {2018}}%
}]{%
dwivedi2018hot}
\APACinsertmetastar {%
dwivedi2018hot}%
\begin{APACrefauthors}%
Dwivedi, D.%
, Arora, B.%
, Steefel, C\BPBI I.%
, Dafflon, B.%
\BCBL {}\ \BBA {} Versteeg, R.%
\end{APACrefauthors}%
\unskip\
\newblock
\APACrefYearMonthDay{2018}{}{}.
\newblock
{\BBOQ}\APACrefatitle {Hot spots and hot moments of nitrogen in a riparian corridor} {Hot spots and hot moments of nitrogen in a riparian corridor}.{\BBCQ}
\newblock
\APACjournalVolNumPages{Water Resources Research}{54}{1}{205--222}.
\PrintBackRefs{\CurrentBib}

\bibitem [\protect \citeauthoryear {%
Ebong%
, Akpan%
\BCBL {}\ \BBA {} Ekwok%
}{%
Ebong%
\ \protect \BOthers {.}}{%
{\protect \APACyear {2020}}%
}]{%
ebong2020stochastic}
\APACinsertmetastar {%
ebong2020stochastic}%
\begin{APACrefauthors}%
Ebong, E\BPBI D.%
, Akpan, A\BPBI E.%
\BCBL {}\ \BBA {} Ekwok, S\BPBI E.%
\end{APACrefauthors}%
\unskip\
\newblock
\APACrefYearMonthDay{2020}{}{}.
\newblock
{\BBOQ}\APACrefatitle {Stochastic modelling of spatial variability of petrophysical properties in parts of the {Niger Delta Basin, southern Nigeria}} {Stochastic modelling of spatial variability of petrophysical properties in parts of the {Niger Delta Basin, southern Nigeria}}.{\BBCQ}
\newblock
\APACjournalVolNumPages{Journal of Petroleum Exploration and Production Technology}{10}{}{569--585}.
\PrintBackRefs{\CurrentBib}

\bibitem [\protect \citeauthoryear {%
Emerick%
\ \BBA {} Reynolds%
}{%
Emerick%
\ \BBA {} Reynolds%
}{%
{\protect \APACyear {2013}}%
}]{%
emerick2013ensemble}
\APACinsertmetastar {%
emerick2013ensemble}%
\begin{APACrefauthors}%
Emerick, A\BPBI A.%
\BCBT {}\ \BBA {} Reynolds, A\BPBI C.%
\end{APACrefauthors}%
\unskip\
\newblock
\APACrefYearMonthDay{2013}{}{}.
\newblock
{\BBOQ}\APACrefatitle {Ensemble smoother with multiple data assimilation} {Ensemble smoother with multiple data assimilation}.{\BBCQ}
\newblock
\APACjournalVolNumPages{Computers \& Geosciences}{55}{}{3--15}.
\PrintBackRefs{\CurrentBib}

\bibitem [\protect \citeauthoryear {%
Goebel%
, Pidlisecky%
\BCBL {}\ \BBA {} Knight%
}{%
Goebel%
\ \protect \BOthers {.}}{%
{\protect \APACyear {2017}}%
}]{%
goebel2017resistivity}
\APACinsertmetastar {%
goebel2017resistivity}%
\begin{APACrefauthors}%
Goebel, M.%
, Pidlisecky, A.%
\BCBL {}\ \BBA {} Knight, R.%
\end{APACrefauthors}%
\unskip\
\newblock
\APACrefYearMonthDay{2017}{}{}.
\newblock
{\BBOQ}\APACrefatitle {Resistivity imaging reveals complex pattern of saltwater intrusion along {M}onterey coast} {Resistivity imaging reveals complex pattern of saltwater intrusion along {M}onterey coast}.{\BBCQ}
\newblock
\APACjournalVolNumPages{Journal of Hydrology}{551}{}{746--755}.
\PrintBackRefs{\CurrentBib}

\bibitem [\protect \citeauthoryear {%
Hammond%
, Lichtner%
\BCBL {}\ \BBA {} Mills%
}{%
Hammond%
\ \protect \BOthers {.}}{%
{\protect \APACyear {2014}}%
}]{%
hammond2014evaluating}
\APACinsertmetastar {%
hammond2014evaluating}%
\begin{APACrefauthors}%
Hammond, G\BPBI E.%
, Lichtner, P\BPBI C.%
\BCBL {}\ \BBA {} Mills, R.%
\end{APACrefauthors}%
\unskip\
\newblock
\APACrefYearMonthDay{2014}{}{}.
\newblock
{\BBOQ}\APACrefatitle {Evaluating the performance of parallel subsurface simulators: {An} illustrative example with {PFLOTRAN}} {Evaluating the performance of parallel subsurface simulators: {An} illustrative example with {PFLOTRAN}}.{\BBCQ}
\newblock
\APACjournalVolNumPages{Water Resources Research}{50}{1}{208--228}.
\PrintBackRefs{\CurrentBib}

\bibitem [\protect \citeauthoryear {%
Han%
, Hamon%
, Jiang%
\BCBL {}\ \BBA {} Durlofsky%
}{%
Han%
\ \protect \BOthers {.}}{%
{\protect \APACyear {2024}}%
}]{%
han2024surrogate}
\APACinsertmetastar {%
han2024surrogate}%
\begin{APACrefauthors}%
Han, Y.%
, Hamon, F\BPBI P.%
, Jiang, S.%
\BCBL {}\ \BBA {} Durlofsky, L\BPBI J.%
\end{APACrefauthors}%
\unskip\
\newblock
\APACrefYearMonthDay{2024}{}{}.
\newblock
{\BBOQ}\APACrefatitle {Surrogate model for geological {CO$_2$} storage and its use in hierarchical {MCMC} history matching} {Surrogate model for geological {CO$_2$} storage and its use in hierarchical {MCMC} history matching}.{\BBCQ}
\newblock
\APACjournalVolNumPages{Advances in Water Resources}{187}{}{104678}.
\PrintBackRefs{\CurrentBib}

\bibitem [\protect \citeauthoryear {%
He%
, H{\o}jberg%
, J{\o}rgensen%
\BCBL {}\ \BBA {} Refsgaard%
}{%
He%
\ \protect \BOthers {.}}{%
{\protect \APACyear {2015}}%
}]{%
he2015assessing}
\APACinsertmetastar {%
he2015assessing}%
\begin{APACrefauthors}%
He, X.%
, H{\o}jberg, A\BPBI L.%
, J{\o}rgensen, F.%
\BCBL {}\ \BBA {} Refsgaard, J\BPBI C.%
\end{APACrefauthors}%
\unskip\
\newblock
\APACrefYearMonthDay{2015}{}{}.
\newblock
{\BBOQ}\APACrefatitle {Assessing hydrological model predictive uncertainty using stochastically generated geological models} {Assessing hydrological model predictive uncertainty using stochastically generated geological models}.{\BBCQ}
\newblock
\APACjournalVolNumPages{Hydrological Processes}{29}{19}{4293--4311}.
\PrintBackRefs{\CurrentBib}

\bibitem [\protect \citeauthoryear {%
Hussain%
, Javadi%
, Ahangar-Asr%
\BCBL {}\ \BBA {} Farmani%
}{%
Hussain%
\ \protect \BOthers {.}}{%
{\protect \APACyear {2015}}%
}]{%
hussain2015surrogate}
\APACinsertmetastar {%
hussain2015surrogate}%
\begin{APACrefauthors}%
Hussain, M\BPBI S.%
, Javadi, A\BPBI A.%
, Ahangar-Asr, A.%
\BCBL {}\ \BBA {} Farmani, R.%
\end{APACrefauthors}%
\unskip\
\newblock
\APACrefYearMonthDay{2015}{}{}.
\newblock
{\BBOQ}\APACrefatitle {A surrogate model for simulation--optimization of aquifer systems subjected to seawater intrusion} {A surrogate model for simulation--optimization of aquifer systems subjected to seawater intrusion}.{\BBCQ}
\newblock
\APACjournalVolNumPages{Journal of Hydrology}{523}{}{542--554}.
\PrintBackRefs{\CurrentBib}

\bibitem [\protect \citeauthoryear {%
Jiang%
\ \BBA {} Durlofsky%
}{%
Jiang%
\ \BBA {} Durlofsky%
}{%
{\protect \APACyear {2023}}%
}]{%
jiang2023use}
\APACinsertmetastar {%
jiang2023use}%
\begin{APACrefauthors}%
Jiang, S.%
\BCBT {}\ \BBA {} Durlofsky, L\BPBI J.%
\end{APACrefauthors}%
\unskip\
\newblock
\APACrefYearMonthDay{2023}{}{}.
\newblock
{\BBOQ}\APACrefatitle {Use of multifidelity training data and transfer learning for efficient construction of subsurface flow surrogate models} {Use of multifidelity training data and transfer learning for efficient construction of subsurface flow surrogate models}.{\BBCQ}
\newblock
\APACjournalVolNumPages{Journal of Computational Physics}{474}{}{111800}.
\PrintBackRefs{\CurrentBib}

\bibitem [\protect \citeauthoryear {%
Jiang%
\ \BBA {} Durlofsky%
}{%
Jiang%
\ \BBA {} Durlofsky%
}{%
{\protect \APACyear {2024}}%
}]{%
jiang2024history}
\APACinsertmetastar {%
jiang2024history}%
\begin{APACrefauthors}%
Jiang, S.%
\BCBT {}\ \BBA {} Durlofsky, L\BPBI J.%
\end{APACrefauthors}%
\unskip\
\newblock
\APACrefYearMonthDay{2024}{}{}.
\newblock
{\BBOQ}\APACrefatitle {History matching for geological carbon storage using data-space inversion with spatio-temporal data parameterization} {History matching for geological carbon storage using data-space inversion with spatio-temporal data parameterization}.{\BBCQ}
\newblock
\APACjournalVolNumPages{International Journal of Greenhouse Gas Control}{134}{}{104124}.
\PrintBackRefs{\CurrentBib}

\bibitem [\protect \citeauthoryear {%
Ketabchi%
, Mahmoodzadeh%
, Ataie-Ashtiani%
\BCBL {}\ \BBA {} Simmons%
}{%
Ketabchi%
\ \protect \BOthers {.}}{%
{\protect \APACyear {2016}}%
}]{%
ketabchi2016sea}
\APACinsertmetastar {%
ketabchi2016sea}%
\begin{APACrefauthors}%
Ketabchi, H.%
, Mahmoodzadeh, D.%
, Ataie-Ashtiani, B.%
\BCBL {}\ \BBA {} Simmons, C\BPBI T.%
\end{APACrefauthors}%
\unskip\
\newblock
\APACrefYearMonthDay{2016}{}{}.
\newblock
{\BBOQ}\APACrefatitle {Sea-level rise impacts on seawater intrusion in coastal aquifers: {R}eview and integration} {Sea-level rise impacts on seawater intrusion in coastal aquifers: {R}eview and integration}.{\BBCQ}
\newblock
\APACjournalVolNumPages{Journal of Hydrology}{535}{}{235--255}.
\PrintBackRefs{\CurrentBib}

\bibitem [\protect \citeauthoryear {%
Kingma%
\ \BBA {} Ba%
}{%
Kingma%
\ \BBA {} Ba%
}{%
{\protect \APACyear {2014}}%
}]{%
kingma2014adam}
\APACinsertmetastar {%
kingma2014adam}%
\begin{APACrefauthors}%
Kingma, D\BPBI P.%
\BCBT {}\ \BBA {} Ba, J.%
\end{APACrefauthors}%
\unskip\
\newblock
\APACrefYearMonthDay{2014}{}{}.
\newblock
{\BBOQ}\APACrefatitle {{A}dam: {A} method for stochastic optimization} {{A}dam: {A} method for stochastic optimization}.{\BBCQ}
\newblock
\APACjournalVolNumPages{arXiv preprint arXiv:1412.6980}{}{}{}.
\PrintBackRefs{\CurrentBib}

\bibitem [\protect \citeauthoryear {%
Langevin%
\ \BBA {} Guo%
}{%
Langevin%
\ \BBA {} Guo%
}{%
{\protect \APACyear {2006}}%
}]{%
langevin2006modflow}
\APACinsertmetastar {%
langevin2006modflow}%
\begin{APACrefauthors}%
Langevin, C\BPBI D.%
\BCBT {}\ \BBA {} Guo, W.%
\end{APACrefauthors}%
\unskip\
\newblock
\APACrefYearMonthDay{2006}{}{}.
\newblock
{\BBOQ}\APACrefatitle {{MODFLOW/MT3DMS}--based simulation of variable-density ground water flow and transport} {{MODFLOW/MT3DMS}--based simulation of variable-density ground water flow and transport}.{\BBCQ}
\newblock
\APACjournalVolNumPages{Groundwater}{44}{3}{339--351}.
\PrintBackRefs{\CurrentBib}

\bibitem [\protect \citeauthoryear {%
Langevin%
, Thorne~Jr%
, Dausman%
, Sukop%
\BCBL {}\ \BBA {} Guo%
}{%
Langevin%
\ \protect \BOthers {.}}{%
{\protect \APACyear {2008}}%
}]{%
langevin2008seawat}
\APACinsertmetastar {%
langevin2008seawat}%
\begin{APACrefauthors}%
Langevin, C\BPBI D.%
, Thorne~Jr, D\BPBI T.%
, Dausman, A\BPBI M.%
, Sukop, M\BPBI C.%
\BCBL {}\ \BBA {} Guo, W.%
\end{APACrefauthors}%
\unskip\
\newblock
\APACrefYearMonthDay{2008}{}{}.
\newblock
\APACrefbtitle {{SEAWAT} version 4: a computer program for simulation of multi-species solute and heat transport} {{SEAWAT} version 4: a computer program for simulation of multi-species solute and heat transport}\ \APACbVolEdTR{}{\BTR{}}.
\newblock
\APACaddressInstitution{}{Geological Survey (US)}.
\PrintBackRefs{\CurrentBib}

\bibitem [\protect \citeauthoryear {%
Li%
\ \protect \BOthers {.}}{%
Li%
\ \protect \BOthers {.}}{%
{\protect \APACyear {2020}}%
}]{%
li2020fourier}
\APACinsertmetastar {%
li2020fourier}%
\begin{APACrefauthors}%
Li, Z.%
, Kovachki, N.%
, Azizzadenesheli, K.%
, Liu, B.%
, Bhattacharya, K.%
, Stuart, A.%
\BCBL {}\ \BBA {} Anandkumar, A.%
\end{APACrefauthors}%
\unskip\
\newblock
\APACrefYearMonthDay{2020}{}{}.
\newblock
{\BBOQ}\APACrefatitle {Fourier neural operator for parametric partial differential equations} {Fourier neural operator for parametric partial differential equations}.{\BBCQ}
\newblock
\APACjournalVolNumPages{arXiv preprint arXiv:2010.08895}{}{}{}.
\PrintBackRefs{\CurrentBib}

\bibitem [\protect \citeauthoryear {%
Meray%
\ \protect \BOthers {.}}{%
Meray%
\ \protect \BOthers {.}}{%
{\protect \APACyear {2024}}%
}]{%
meray2024physics}
\APACinsertmetastar {%
meray2024physics}%
\begin{APACrefauthors}%
Meray, A.%
, Wang, L.%
, Kurihana, T.%
, Mastilovic, I.%
, Praveen, S.%
, Xu, Z.%
\BDBL {}Wainwright, H.%
\end{APACrefauthors}%
\unskip\
\newblock
\APACrefYearMonthDay{2024}{}{}.
\newblock
{\BBOQ}\APACrefatitle {Physics-informed surrogate modeling for supporting climate resilience at groundwater contamination sites} {Physics-informed surrogate modeling for supporting climate resilience at groundwater contamination sites}.{\BBCQ}
\newblock
\APACjournalVolNumPages{Computers \& Geosciences}{183}{}{105508}.
\PrintBackRefs{\CurrentBib}

\bibitem [\protect \citeauthoryear {%
Messier%
, Campbell%
, Bradley%
\BCBL {}\ \BBA {} Serre%
}{%
Messier%
\ \protect \BOthers {.}}{%
{\protect \APACyear {2015}}%
}]{%
messier2015estimation}
\APACinsertmetastar {%
messier2015estimation}%
\begin{APACrefauthors}%
Messier, K\BPBI P.%
, Campbell, T.%
, Bradley, P\BPBI J.%
\BCBL {}\ \BBA {} Serre, M\BPBI L.%
\end{APACrefauthors}%
\unskip\
\newblock
\APACrefYearMonthDay{2015}{}{}.
\newblock
{\BBOQ}\APACrefatitle {Estimation of groundwater {Radon in North Carolina} using land use regression and {B}ayesian maximum entropy} {Estimation of groundwater {Radon in North Carolina} using land use regression and {B}ayesian maximum entropy}.{\BBCQ}
\newblock
\APACjournalVolNumPages{Environmental Science \& Technology}{49}{16}{9817--9825}.
\PrintBackRefs{\CurrentBib}

\bibitem [\protect \citeauthoryear {%
Mo%
, Zabaras%
, Shi%
\BCBL {}\ \BBA {} Wu%
}{%
Mo%
\ \protect \BOthers {.}}{%
{\protect \APACyear {2019}}%
}]{%
mo2019deep}
\APACinsertmetastar {%
mo2019deep}%
\begin{APACrefauthors}%
Mo, S.%
, Zabaras, N.%
, Shi, X.%
\BCBL {}\ \BBA {} Wu, J.%
\end{APACrefauthors}%
\unskip\
\newblock
\APACrefYearMonthDay{2019}{}{}.
\newblock
{\BBOQ}\APACrefatitle {Deep autoregressive neural networks for high-dimensional inverse problems in groundwater contaminant source identification} {Deep autoregressive neural networks for high-dimensional inverse problems in groundwater contaminant source identification}.{\BBCQ}
\newblock
\APACjournalVolNumPages{Water Resources Research}{55}{5}{3856--3881}.
\PrintBackRefs{\CurrentBib}

\bibitem [\protect \citeauthoryear {%
Provost%
\ \BBA {} Voss%
}{%
Provost%
\ \BBA {} Voss%
}{%
{\protect \APACyear {2019}}%
}]{%
provost2019sutra}
\APACinsertmetastar {%
provost2019sutra}%
\begin{APACrefauthors}%
Provost, A\BPBI M.%
\BCBT {}\ \BBA {} Voss, C\BPBI I.%
\end{APACrefauthors}%
\unskip\
\newblock
\APACrefYearMonthDay{2019}{}{}.
\newblock
\APACrefbtitle {{SUTRA}, a model for saturated-unsaturated, variable-density groundwater flow with solute or energy transport—Documentation of generalized boundary conditions, a modified implementation of specified pressures and concentrations or temperatures, and the lake capability} {{SUTRA}, a model for saturated-unsaturated, variable-density groundwater flow with solute or energy transport—documentation of generalized boundary conditions, a modified implementation of specified pressures and concentrations or temperatures, and the lake capability}\ \APACbVolEdTR{}{\BTR{}}.
\newblock
\APACaddressInstitution{}{US Geological Survey}.
\PrintBackRefs{\CurrentBib}

\bibitem [\protect \citeauthoryear {%
Rajabi%
\ \BBA {} Ketabchi%
}{%
Rajabi%
\ \BBA {} Ketabchi%
}{%
{\protect \APACyear {2017}}%
}]{%
rajabi2017uncertainty}
\APACinsertmetastar {%
rajabi2017uncertainty}%
\begin{APACrefauthors}%
Rajabi, M\BPBI M.%
\BCBT {}\ \BBA {} Ketabchi, H.%
\end{APACrefauthors}%
\unskip\
\newblock
\APACrefYearMonthDay{2017}{}{}.
\newblock
{\BBOQ}\APACrefatitle {Uncertainty-based simulation-optimization using {G}aussian process emulation: application to coastal groundwater management} {Uncertainty-based simulation-optimization using {G}aussian process emulation: application to coastal groundwater management}.{\BBCQ}
\newblock
\APACjournalVolNumPages{Journal of Hydrology}{555}{}{518--534}.
\PrintBackRefs{\CurrentBib}

\bibitem [\protect \citeauthoryear {%
Remy%
, Boucher%
\BCBL {}\ \BBA {} Wu%
}{%
Remy%
\ \protect \BOthers {.}}{%
{\protect \APACyear {2009}}%
}]{%
remy2009applied}
\APACinsertmetastar {%
remy2009applied}%
\begin{APACrefauthors}%
Remy, N.%
, Boucher, A.%
\BCBL {}\ \BBA {} Wu, J.%
\end{APACrefauthors}%
\unskip\
\newblock
\APACrefYear{2009}.
\newblock
\APACrefbtitle {{Applied geostatistics with {SGeMS}: {A} user's guide}} {{Applied geostatistics with {SGeMS}: {A} user's guide}}.
\newblock
\APACaddressPublisher{}{Cambridge University Press}.
\PrintBackRefs{\CurrentBib}

\bibitem [\protect \citeauthoryear {%
Siler%
\ \protect \BOthers {.}}{%
Siler%
\ \protect \BOthers {.}}{%
{\protect \APACyear {2019}}%
}]{%
siler2019three}
\APACinsertmetastar {%
siler2019three}%
\begin{APACrefauthors}%
Siler, D\BPBI L.%
, Faulds, J\BPBI E.%
, Hinz, N\BPBI H.%
, Dering, G\BPBI M.%
, Edwards, J\BPBI H.%
\BCBL {}\ \BBA {} Mayhew, B.%
\end{APACrefauthors}%
\unskip\
\newblock
\APACrefYearMonthDay{2019}{}{}.
\newblock
{\BBOQ}\APACrefatitle {Three-dimensional geologic mapping to assess geothermal potential: {E}xamples from {N}evada and {O}regon} {Three-dimensional geologic mapping to assess geothermal potential: {E}xamples from {N}evada and {O}regon}.{\BBCQ}
\newblock
\APACjournalVolNumPages{Geothermal Energy}{7}{}{1--32}.
\PrintBackRefs{\CurrentBib}

\bibitem [\protect \citeauthoryear {%
Sonnenborg%
, Seifert%
\BCBL {}\ \BBA {} Refsgaard%
}{%
Sonnenborg%
\ \protect \BOthers {.}}{%
{\protect \APACyear {2015}}%
}]{%
sonnenborg2015climate}
\APACinsertmetastar {%
sonnenborg2015climate}%
\begin{APACrefauthors}%
Sonnenborg, T.%
, Seifert, D.%
\BCBL {}\ \BBA {} Refsgaard, J.%
\end{APACrefauthors}%
\unskip\
\newblock
\APACrefYearMonthDay{2015}{}{}.
\newblock
{\BBOQ}\APACrefatitle {Climate model uncertainty versus conceptual geological uncertainty in hydrological modeling} {Climate model uncertainty versus conceptual geological uncertainty in hydrological modeling}.{\BBCQ}
\newblock
\APACjournalVolNumPages{Hydrology and Earth System Sciences}{19}{9}{3891--3901}.
\PrintBackRefs{\CurrentBib}

\bibitem [\protect \citeauthoryear {%
Sreekanth%
\ \BBA {} Datta%
}{%
Sreekanth%
\ \BBA {} Datta%
}{%
{\protect \APACyear {2010}}%
}]{%
sreekanth2010multi}
\APACinsertmetastar {%
sreekanth2010multi}%
\begin{APACrefauthors}%
Sreekanth, J.%
\BCBT {}\ \BBA {} Datta, B.%
\end{APACrefauthors}%
\unskip\
\newblock
\APACrefYearMonthDay{2010}{}{}.
\newblock
{\BBOQ}\APACrefatitle {Multi-objective management of saltwater intrusion in coastal aquifers using genetic programming and modular neural network based surrogate models} {Multi-objective management of saltwater intrusion in coastal aquifers using genetic programming and modular neural network based surrogate models}.{\BBCQ}
\newblock
\APACjournalVolNumPages{Journal of Hydrology}{393}{3-4}{245--256}.
\PrintBackRefs{\CurrentBib}

\bibitem [\protect \citeauthoryear {%
Strati%
, Wipperfurth%
, Baldoncini%
, McDonough%
\BCBL {}\ \BBA {} Mantovani%
}{%
Strati%
\ \protect \BOthers {.}}{%
{\protect \APACyear {2017}}%
}]{%
strati2017perceiving}
\APACinsertmetastar {%
strati2017perceiving}%
\begin{APACrefauthors}%
Strati, V.%
, Wipperfurth, S\BPBI A.%
, Baldoncini, M.%
, McDonough, W\BPBI F.%
\BCBL {}\ \BBA {} Mantovani, F.%
\end{APACrefauthors}%
\unskip\
\newblock
\APACrefYearMonthDay{2017}{}{}.
\newblock
{\BBOQ}\APACrefatitle {Perceiving the crust in 3-{D}: {A} model integrating geological, geochemical, and geophysical data} {Perceiving the crust in 3-{D}: {A} model integrating geological, geochemical, and geophysical data}.{\BBCQ}
\newblock
\APACjournalVolNumPages{Geochemistry, Geophysics, Geosystems}{18}{12}{4326--4341}.
\PrintBackRefs{\CurrentBib}

\bibitem [\protect \citeauthoryear {%
Tang%
, Liu%
\BCBL {}\ \BBA {} Durlofsky%
}{%
Tang%
\ \protect \BOthers {.}}{%
{\protect \APACyear {2020}}%
}]{%
tang2020deep}
\APACinsertmetastar {%
tang2020deep}%
\begin{APACrefauthors}%
Tang, M.%
, Liu, Y.%
\BCBL {}\ \BBA {} Durlofsky, L\BPBI J.%
\end{APACrefauthors}%
\unskip\
\newblock
\APACrefYearMonthDay{2020}{}{}.
\newblock
{\BBOQ}\APACrefatitle {A deep-learning-based surrogate model for data assimilation in dynamic subsurface flow problems} {A deep-learning-based surrogate model for data assimilation in dynamic subsurface flow problems}.{\BBCQ}
\newblock
\APACjournalVolNumPages{Journal of Computational Physics}{413}{}{109456}.
\PrintBackRefs{\CurrentBib}

\bibitem [\protect \citeauthoryear {%
Torresan%
, Piccinini%
, Pola%
, Zampieri%
\BCBL {}\ \BBA {} Fabbri%
}{%
Torresan%
\ \protect \BOthers {.}}{%
{\protect \APACyear {2020}}%
}]{%
torresan20203d}
\APACinsertmetastar {%
torresan20203d}%
\begin{APACrefauthors}%
Torresan, F.%
, Piccinini, L.%
, Pola, M.%
, Zampieri, D.%
\BCBL {}\ \BBA {} Fabbri, P.%
\end{APACrefauthors}%
\unskip\
\newblock
\APACrefYearMonthDay{2020}{}{}.
\newblock
{\BBOQ}\APACrefatitle {3{D} hydrogeological reconstruction of the fault-controlled {Euganean Geothermal System (NE Italy)}} {3{D} hydrogeological reconstruction of the fault-controlled {Euganean Geothermal System (NE Italy)}}.{\BBCQ}
\newblock
\APACjournalVolNumPages{Engineering Geology}{274}{}{105740}.
\PrintBackRefs{\CurrentBib}

\bibitem [\protect \citeauthoryear {%
Wen%
, Li%
, Azizzadenesheli%
, Anandkumar%
\BCBL {}\ \BBA {} Benson%
}{%
Wen%
\ \protect \BOthers {.}}{%
{\protect \APACyear {2022}}%
}]{%
wen2022u}
\APACinsertmetastar {%
wen2022u}%
\begin{APACrefauthors}%
Wen, G.%
, Li, Z.%
, Azizzadenesheli, K.%
, Anandkumar, A.%
\BCBL {}\ \BBA {} Benson, S\BPBI M.%
\end{APACrefauthors}%
\unskip\
\newblock
\APACrefYearMonthDay{2022}{}{}.
\newblock
{\BBOQ}\APACrefatitle {{U-FNO}—An enhanced {F}ourier neural operator-based deep-learning model for multiphase flow} {{U-FNO}—an enhanced {F}ourier neural operator-based deep-learning model for multiphase flow}.{\BBCQ}
\newblock
\APACjournalVolNumPages{Advances in Water Resources}{163}{}{104180}.
\PrintBackRefs{\CurrentBib}

\bibitem [\protect \citeauthoryear {%
Werner%
\ \protect \BOthers {.}}{%
Werner%
\ \protect \BOthers {.}}{%
{\protect \APACyear {2013}}%
}]{%
werner2013seawater}
\APACinsertmetastar {%
werner2013seawater}%
\begin{APACrefauthors}%
Werner, A\BPBI D.%
, Bakker, M.%
, Post, V\BPBI E.%
, Vandenbohede, A.%
, Lu, C.%
, Ataie-Ashtiani, B.%
\BDBL {}Barry, D\BPBI A.%
\end{APACrefauthors}%
\unskip\
\newblock
\APACrefYearMonthDay{2013}{}{}.
\newblock
{\BBOQ}\APACrefatitle {Seawater intrusion processes, investigation and management: {R}ecent advances and future challenges} {Seawater intrusion processes, investigation and management: {R}ecent advances and future challenges}.{\BBCQ}
\newblock
\APACjournalVolNumPages{Advances in Water Resources}{51}{}{3--26}.
\PrintBackRefs{\CurrentBib}

\bibitem [\protect \citeauthoryear {%
Yabusaki%
\ \protect \BOthers {.}}{%
Yabusaki%
\ \protect \BOthers {.}}{%
{\protect \APACyear {2020}}%
}]{%
yabusaki2020floodplain}
\APACinsertmetastar {%
yabusaki2020floodplain}%
\begin{APACrefauthors}%
Yabusaki, S\BPBI B.%
, Myers-Pigg, A\BPBI N.%
, Ward, N\BPBI D.%
, Waichler, S\BPBI R.%
, Sengupta, A.%
, Hou, Z.%
\BDBL {}others%
\end{APACrefauthors}%
\unskip\
\newblock
\APACrefYearMonthDay{2020}{}{}.
\newblock
{\BBOQ}\APACrefatitle {Floodplain inundation and salinization from a recently restored first-order tidal stream} {Floodplain inundation and salinization from a recently restored first-order tidal stream}.{\BBCQ}
\newblock
\APACjournalVolNumPages{Water Resources Research}{56}{7}{e2019WR026850}.
\PrintBackRefs{\CurrentBib}

\bibitem [\protect \citeauthoryear {%
Yoon%
, Williams%
, Juanes%
\BCBL {}\ \BBA {} Kang%
}{%
Yoon%
\ \protect \BOthers {.}}{%
{\protect \APACyear {2017}}%
}]{%
yoon2017maximizing}
\APACinsertmetastar {%
yoon2017maximizing}%
\begin{APACrefauthors}%
Yoon, S.%
, Williams, J\BPBI R.%
, Juanes, R.%
\BCBL {}\ \BBA {} Kang, P\BPBI K.%
\end{APACrefauthors}%
\unskip\
\newblock
\APACrefYearMonthDay{2017}{}{}.
\newblock
{\BBOQ}\APACrefatitle {Maximizing the value of pressure data in saline aquifer characterization} {Maximizing the value of pressure data in saline aquifer characterization}.{\BBCQ}
\newblock
\APACjournalVolNumPages{Advances in Water Resources}{109}{}{14--28}.
\PrintBackRefs{\CurrentBib}

\bibitem [\protect \citeauthoryear {%
Zhong%
, Sun%
\BCBL {}\ \BBA {} Jeong%
}{%
Zhong%
\ \protect \BOthers {.}}{%
{\protect \APACyear {2019}}%
}]{%
zhong2019predicting}
\APACinsertmetastar {%
zhong2019predicting}%
\begin{APACrefauthors}%
Zhong, Z.%
, Sun, A\BPBI Y.%
\BCBL {}\ \BBA {} Jeong, H.%
\end{APACrefauthors}%
\unskip\
\newblock
\APACrefYearMonthDay{2019}{}{}.
\newblock
{\BBOQ}\APACrefatitle {Predicting {CO$_2$} plume migration in heterogeneous formations using conditional deep convolutional generative adversarial network} {Predicting {CO$_2$} plume migration in heterogeneous formations using conditional deep convolutional generative adversarial network}.{\BBCQ}
\newblock
\APACjournalVolNumPages{Water Resources Research}{55}{7}{5830--5851}.
\PrintBackRefs{\CurrentBib}

\bibitem [\protect \citeauthoryear {%
Zhou%
\ \BBA {} Tartakovsky%
}{%
Zhou%
\ \BBA {} Tartakovsky%
}{%
{\protect \APACyear {2021}}%
}]{%
zhou2021markov}
\APACinsertmetastar {%
zhou2021markov}%
\begin{APACrefauthors}%
Zhou, Z.%
\BCBT {}\ \BBA {} Tartakovsky, D\BPBI M.%
\end{APACrefauthors}%
\unskip\
\newblock
\APACrefYearMonthDay{2021}{}{}.
\newblock
{\BBOQ}\APACrefatitle {{Markov chain Monte Carlo} with neural network surrogates: {A}pplication to contaminant source identification} {{Markov chain Monte Carlo} with neural network surrogates: {A}pplication to contaminant source identification}.{\BBCQ}
\newblock
\APACjournalVolNumPages{Stochastic Environmental Research and Risk Assessment}{35}{}{639--651}.
\PrintBackRefs{\CurrentBib}

\bibitem [\protect \citeauthoryear {%
Zhou%
, Zabaras%
\BCBL {}\ \BBA {} Tartakovsky%
}{%
Zhou%
\ \protect \BOthers {.}}{%
{\protect \APACyear {2022}}%
}]{%
zhou2022deep}
\APACinsertmetastar {%
zhou2022deep}%
\begin{APACrefauthors}%
Zhou, Z.%
, Zabaras, N.%
\BCBL {}\ \BBA {} Tartakovsky, D\BPBI M.%
\end{APACrefauthors}%
\unskip\
\newblock
\APACrefYearMonthDay{2022}{}{}.
\newblock
{\BBOQ}\APACrefatitle {Deep learning for simultaneous inference of hydraulic and transport properties} {Deep learning for simultaneous inference of hydraulic and transport properties}.{\BBCQ}
\newblock
\APACjournalVolNumPages{Water Resources Research}{58}{10}{e2021WR031438}.
\PrintBackRefs{\CurrentBib}

\end{thebibliography}
